\documentclass[lettersize,journal]{IEEEtran}
\usepackage{amsmath,amsfonts}
\usepackage{algorithmic}
\usepackage{algorithm}
\usepackage{array}
\usepackage{amsmath}
\usepackage{float} 
\usepackage{hyperref}
\usepackage[caption=false,font=normalsize,labelfont=sf,textfont=sf]{subfig}
\usepackage{textcomp}
\usepackage{stfloats}
\usepackage{url}
\usepackage{verbatim}
\usepackage{graphicx}
\usepackage{cite}
\usepackage{xcolor}
\usepackage{booktabs}
\usepackage{multirow, makecell}
\usepackage{gensymb}
\usepackage{mdframed}
\usepackage{amssymb}
\usepackage{pifont}
\newcommand{\cmark}{\ding{51}}%
\newcommand{\xmark}{\ding{55}}%
\definecolor{darkgreen}{RGB}{0, 176, 80} 

\begin{document}

\title{Robust Change Captioning in Remote Sensing: SECOND-CC Dataset and MModalCC Framework}

\author{Ali Can Karaca, M. Enes Özelbaş, Saadettin Berber, Orkhan Karimli, Turabi Yildirim,  M. Fatih Amasyali %
\thanks{A. C. Karaca, E. Ozelbas, S. Berber, O. Karimli, T. Yildirim, and M. F. Amasyali are with the Department of Computer Engineering, Yildiz Technical University, 34220 Istanbul, Turkey (e-mail:  ackaraca@yildiz.edu.tr; enes.ozelbas@yildiz.edu.tr; saadettin.berber@std.yildiz.edu.tr; orkhan.karimli@std.yildiz.edu.tr; turabi.yildirim@std.yildiz.edu.tr; amasyali@yildiz.edu.tr). The authors also with the Multimodal Systems for Artificial Intelligence and Computation (MOSAIC) Research Group, Yildiz Technical University, 34220 Istanbul, Turkey. This research was supported by the Scientific and Technological Research Council of Turkey (TUBITAK) under grant 3501, project number 122E666.}}

\markboth{Submitted to IEEE TRANSACTIONS ON GEOSCIENCE AND REMOTE SENSING}%
{Berber \MakeLowercase{\textit{et al.}}: SECOND-CC: Remote Sensing Change Captioning
with Semantic Maps}


\maketitle

\begin{abstract}
Remote sensing change captioning (RSICC) aims to describe changes between bitemporal images in natural language. Existing methods often fail under challenges like illumination differences, viewpoint changes, blur effects, leading to inaccuracies, especially in no-change regions. Moreover, the images acquired at different spatial resolutions and have registration errors tend to affect the captions. To address these issues, we introduce SECOND-CC, a novel RSICC dataset featuring high-resolution RGB image pairs, semantic segmentation maps, and diverse real-world scenarios. SECOND-CC which contains 6\,041 pairs of bitemporal RS images and 30\,205 sentences describing the differences between images. Additionally, we propose MModalCC, a multimodal framework that integrates semantic and visual data using advanced attention mechanisms, including Cross-Modal Cross Attention (CMCA) and Multimodal Gated Cross Attention (MGCA). Detailed ablation studies and attention visualizations further demonstrate its effectiveness and ability to address RSICC challenges. Comprehensive experiments show that MModalCC outperforms state-of-the-art RSICC methods, including RSICCformer, Chg2Cap, and PSNet with +4.6\% improvement on BLEU4 score and +9.6\% improvement on CIDEr score. We will make our dataset and codebase publicly available to facilitate future research at \href{https://github.com/ChangeCapsInRS/SecondCC}{https://github.com/ChangeCapsInRS/SecondCC}.
\end{abstract}


\section{Introduction}

Remote Sensing Image Change Captioning (RSICC) has emerged as a pivotal field of research for providing meaningful natural language descriptions of detected changes in raw bi-temporal remote sensing imagery~\cite{hoxha2022change,li2024inter}. Unlike traditional change detection tasks, which primarily provide classes per pixel, RSICC interprets changes contextually, providing detailed information on objects, their spatial transformations, and the temporal relations involved. Accurate change interpretations become vital for many fields including disaster response, urban planning, environmental monitoring, and security, enabling descriptive insights for damage control, land use, environmental changes, and infrastructure changes for timely and informed action. RSICC frameworks typically consist of a vision encoder to extract visual features from multitemporal images and a caption decoder to translate these features into captions that describe spatial changes in the context of temporal evolution.

Early works, such as RSICCformer~\cite{RSICCformer-LEVIR-CC}, laid the groundwork for remote sensing change captioning (RSICC) by introducing a dual-branch Transformer encoder. This architecture extracted high-level semantic features and leveraged bitemporal feature differences to improve change discrimination, establishing the LEVIR-CC dataset as a widely used benchmark in the field. PSNet~\cite{liu2023progressivescaleawarenetworkremote} incorporated multi-scale feature extraction through progressive difference perception (PDP) layers and scale-aware reinforcement (SR) modules, focusing on identifying changes across varying object scales. MAF-Net~\cite{Chen_2024} extended this approach by proposing Multi-Scale Change Aware Encoders (MCAE) with a Gated Attentive Fusion (GAF) module to aggregate change-aware features across multiple scales. Moreover, ICT-Net~\cite{Cai_2023} introduced the Interactive Change-Aware Encoder (ICE) and Multi-Layer Adaptive Fusion (MAF) modules to selectively extract discriminative features while minimizing irrelevant information. Additionally, ICT-Net introduced the Cross Gated-Attention (CGA) module, which integrates multi-scale change-aware features to enhance captioning performance. 

Recently, Semantic-CC~\cite{zhu2024semantic} and Change-Agent~\cite{Liu_2024} have demonstrated innovative approaches by incorporating multimodal feature integration with foundation models using Large Language Models (LLMs). Semantic-CC employs the ViT-based Segment Anything Model (SAM) for feature extraction and the Vicuna-7B LLM for decoding. Change-Agent, on the other hand, introduces a dual-branch Multi-task Change Interpretation (MCI) model supported by an LLM for user interaction and additional analyses~\cite{Liu_2024}. While Semantic-CC focuses on balancing semantic change understanding and detection to present a change captioning model in a traditional sense, Change-Agent expands functionality to include object counting and predictive insights with chatting tools for prompt-based user interaction. Although a large number of researches have been developed over the last two years~\cite{EarthGPT, CCExpert}, these models have an extreme number of learnable parameters.

While significant progress has been achieved in the RSICC domain, several challenges and opportunities remain for this task. Most existing frameworks focus on unimodal feature space (e.g., RGB, MS, SAR)~\cite{RSICCformer-LEVIR-CC,CHG2CAP} and face challenges in distinguishing meaningful changes from complex scenes with subtle or small-scale variations~\cite{huang2021image, Cai_2023}. These difficulties are often caused by limitations in data quality, such as alignment issues, variations in image capture angles, and inconsistencies in resolution or lighting, which can obscure critical changes and introduce noise into the analysis. Additionally, establishing complex relationships between objects in image pairs is hindered by missing defining details, which limits the model to generate coherent and contextually accurate captions for intricate scenes. The recent advancements in LLMs have enabled the development of innovative methods that incorporate multimodal feature integration~\cite{zhu2024semantic} and multiprompt strategies~\cite{Liu_2024}. However, many of these approaches rely on computationally intensive, hard-to-balance architectures, limiting their scalability and applicability in real-world, time-sensitive scenarios. Furthermore, these models often demand large datasets to achieve their full potential, reducing their effectiveness when applied to smaller or domain-specific datasets. Addressing these issues necessitates a shift towards lightweight and efficient architectures that can achieve similar levels of captioning performance while reducing computational overhead. Integrating richer contextual information, such as semantic maps, and leveraging efficient cross-modal feature fusion techniques could bridge the gap between performance and scalability, paving the way for broader applicability of RSICC methods.

Remote sensing change captioning (RSICC) faces critical challenges such as differences in illumination intensity, variations in viewpoints, blur effects, seasonal variations, and image misregistration. Moreover, images captured at different spatial resolutions and those with registration errors can further impact the accuracy of the captions. These issues often mislead RSICC methods, resulting in the generation of incorrect captions. In this paper, we introduce a publicly available RSICC dataset, SECOND-CC, which is specifically designed to address these challenges. Unlike the widely used LEVIR-CC dataset, SECOND-CC incorporates diverse scenarios impacted by these challenges. The dataset includes both color images and semantic segmentation maps, which have been previously utilized for change detection tasks~\cite{SECOND, SECOND2}. To produce detailed change captions under the aforementioned challenges, we propose a novel RSICC method, MModalCC (Multi-Modal Change Captioning), which effectively utilizes multimodal data, including color images and semantic segmentation maps. Our primary contributions are outlined as follows:

\begin{figure*} [hb]
    \centering
    \includegraphics[width=0.80\linewidth]{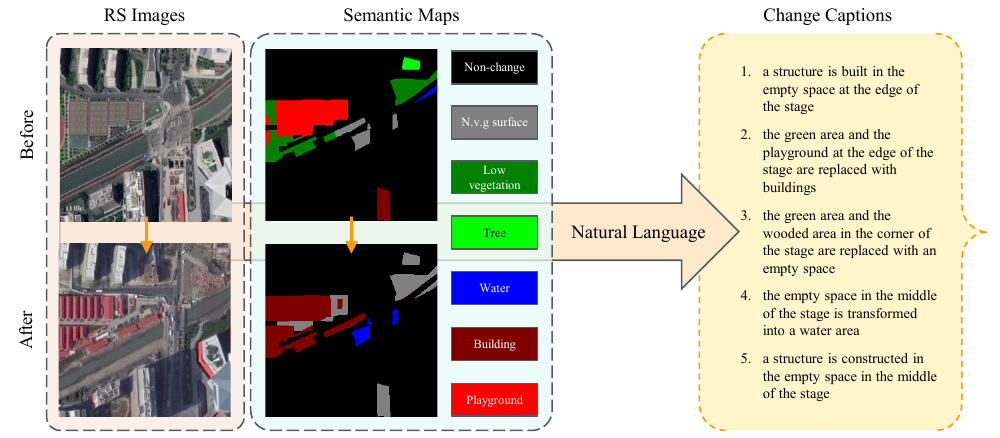}
    \caption{An Example from the SECOND-CC Dataset including color images acquired from two different dates, semantic maps for them, and five change captions that describe the image pair.}
    \label{fig:enter-label}
\end{figure*}

\begin{itemize}
    \item We propose a new multimodal RSICC dataset, SecondCC, tailored for change captioning tasks in remote sensing. This dataset includes high-resolution RGB image pairs, detailed textual descriptions, and semantic maps, enabling comprehensive multi-modal learning. Notably, the images are designed to reflect real-world conditions, including challenges such as blur, intensity and contrast differences, viewpoint variations, and image misregistration.
    \item We design an innovative architecture that efficiently integrates information from semantic segmentation maps and RGB color images by simultaneously capturing cross-modal and unimodal relationships. The architecture includes two siamese encoder for feature extractor, Cross-Modal Cross Attention (CMCA), Unimodal Difference Cross Attention (UDCA), and Convolutional Block (CB) for feature enhancement, and Multimodal Gated Cross Attention (MGCA) based caption decoder.
     \item We perform a detailed comparison and analysis of attention mechanisms in both the image feature representation and caption generation stages. Through systematic analysis and the evaluation of various network configurations, we offer valuable insights to guide researchers in designing multimodal change captioning tasks.
\end{itemize}

\section{SECOND-CC Dataset}
The SECOND (SEmantic Change detectiON Dataset) dataset, a well-known comprehensive remote sensing dataset, contains bitemporal RS images and semantic maps and is designed to address challenges in change detection tasks~\cite{SECOND, SECOND2} and several research studies have been undertaken on this task~\cite{chen2024changemamba,ding2024joint}. In this paper, we propose the SECOND-CC dataset that includes detailed captions for the SECOND change detection dataset. An example from the SECOND-CC dataset can be seen in Figure~\ref{fig:enter-label}.

SECOND-CC expands the contents of its predecessor by adding detailed change captions, which describe the variations between bitemporal remote sensing (RS) image pairs in natural language~\cite{SECOND2}. Moreover, it features labels of the most significant change category, such as \textit{low vegetation} $\rightarrow$ \textit{building}, for the image pairs, enabling comprehensive analysis and customization options.

The lack of high-quality datasets is one of the greatest challenges faced by the RSICC research. The introduction of the SECOND-CC dataset contributes to the advancement of the RSICC task by providing an additional method for training and evaluating an RSICC approach. With the addition of the semantic maps annotating the differences between bitemporal images at the pixel level, it also facilitates change interpretation tasks \cite{LEVIR-MCI}. The proposed dataset can be openly accessed at \href{https://github.com/ChangeCapsInRS/SecondCC}{https://github.com/ChangeCapsInRS/SecondCC}.

\subsection{RS Images}
The RS image pairs in the SECOND dataset capture the aerial views of the cities of Hangzhou, Chengdu, and Shanghai, each with a variety of terrain changes and \textit{distractors}. Compared to the LEVIR-CC dataset \cite{RSICCformer-LEVIR-CC}, the image pairs in SECOND present greater challenges in the form of varying image resolutions, viewpoint change, contrast, luminance, and tonal differences, and alignment issues. Reflecting the complexity of real-world RS applications, these issues serve as a basis for creating a demanding RSICC dataset. 

The SECOND dataset contains RS images that have a size of 512$\times$512, and provide resolutions ranging from 0.5 to 3 meters per pixel. Similar to \cite{RSICCformer-LEVIR-CC}, we divide the images in SECOND equally into four distinct quadrants with dimensions of 256 $\times$ 256. By reducing the amount of change in the image pairs to be annotated, this step results in more manageable chunks for change captioning and increases the amount of available images.

\subsection{Semantic Maps}
The SECOND dataset includes high-quality semantic maps segmenting the changed regions of the scenes. 
The annotations are provided in terms of six different land-cover categories, including \textit{low vegetation}, \textit{non-vegetated ground (n.v.g.) surface}, \textit{tree}, \textit{water}, \textit{building}, and \textit{playground}. The regions of the scene that have not changed between bitemporal states are not labeled with semantic coloring. Notably, these semantic maps can represent multiple categorical changes within a single semantic pair (e.g., low vegetation to building, water to playground). The common transitions between the land-cover categories result in 30 different change categories in SECOND-CC. 

\subsection{Captions}
The changes to the surface of the regions depicted in the RS image pairs are described with five human-labeled sentences. There is a total of 30\,205 sentences in the dataset, prepared with the help of seven contributors in a year.
Through the analysis of the LEVIR-CC and SECOND datasets, several rules and decisions are established to guide the change captioning efforts. Key guidelines include:
\begin{itemize}
    \item \textbf{Detailing Change Characteristics:} It is essential to provide specific information regarding change characteristics, including color, shape, location, intensity, and type, to facilitate a nuanced representation of changes.
    
    \item \textbf{Avoiding Excessive Repetition:} To improve the quality of data generation, it is important to minimize repetitive patterns in captioning.
    
    \item \textbf{Minimizing Emphasis on Peripheral Changes:} During the labeling process, the focus should be limited to significant changes instead of minor edge alterations.
    
    \item \textbf{Stacking the Presentation of Multiple Changes:} Multiple principal alterations should be presented in sequential sentences.
    
    \item \textbf{Dictionary Size Considerations:} The total number of distinct words should be maintained below 2\,000.
    
    \item \textbf{Consistent Use of Directional Phrases:} Directional phrases should be standardized (e.g., use ``top'' instead of ``up'', ``upper'', or ``above''). 
\end{itemize}

\subsection{Data Augmentation}

In this paper, we prepared an augmented dataset SECOND-CC AUG to achieve higher and more stable performances in this challenging data. To this end, we manipulated the images with a hybrid augmentation method that uses well-known transformations such as blurring, brightening, mirroring, and rotating. Note that this augmentation approach was previously proposed for us and recently presented in ~\cite{Augmentation}. Blurring operation is applied with Gaussian blur process with kernel size 5$\times5$ and brightening operation is performed by pixelwise intensity increase in the images. For blurring and brightening, the captions are directly copied. Since mirror and rotate operations change the position and orientation of objects in the images, it is crucial to update the captions to align with their corresponding updated image pairs. The hybrid augmentation method synthesizes a single image pair by randomly choosing one of the related four transformations for each image pair in training and validation sets.

\subsection{Data Analysis}
The SECOND-CC dataset contains 6\,041 data entries and 30\,159 total captions, as stated in Table~\ref{tab:dataset-entry-counts-by-change-status-and-split}. Each entry is composed of a pair of RS images, semantic maps, five captions, and a label that specifies the most significant change on which the annotations are focused. Some of the entries feature RS image pairs with no discernible land-cover change across them, which are referred in the No-Change category. The remaining entries, comprised of the change categories, are denoted under the Change category. According to the table, the number of samples under the Change category constitutes 71.74$\%$ of the entire dataset. The augmented dataset SECOND-CC AUG has twice as many image pairs and sentences as the source dataset, excluding the test split as is seen in Table~\ref{tab:dataset-entry-counts-by-change-status-and-split}. Here, the total number of samples used in SECOND-CC AUG is increased to 10\,855 thanks to the augmentation process.

\begin{table}[htbp]
\centering
\caption{SECOND-CC: Number of Image Pairs for Change and No-Change by Data Split}
\label{tab:dataset-entry-counts-by-change-status-and-split}
\begin{tabular}{lc|rrr|r}
\toprule
\thead[l]{Category} &\thead[r]{Augmentation} & \thead[r]{Train} & \thead[r]{Validation} & \thead[r]{Test} & \thead[r]{Overall} \\ \midrule
No-Change & \xmark & 1\,193 & 170 & 341 & 1\,704 \\
Change & \xmark & 3\,026 & 425 & 886 & 4\,337 \\ 
Total & \xmark & 4\,219 & 595 & 1\,227 & 6\,041 \\ \midrule
No-Change & \cmark & 2\,386 & 340 & 341 & 3\,067 \\
Change & \cmark & 6\,052 & 850 & 886 & 7\,788 \\ 
Total & \cmark & 8\,438 & 1\,190 & 1\,227 & 10\,855 \\
\bottomrule
\end{tabular}
\end{table}

\begin{figure}[htbp]
    \centering
    \includegraphics[width=1\linewidth]{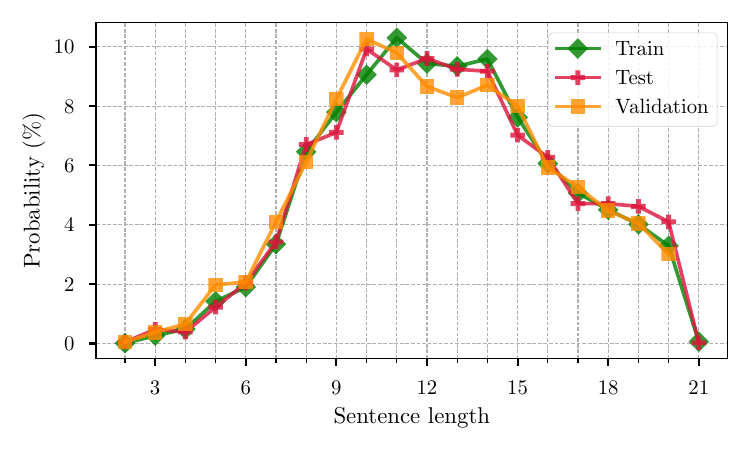}
    \caption{Probability Distribution of Sentence Lengths in Change Data by Data Split}
    \label{fig:change-data-sentence-length-dist}
\end{figure}

The dataset is divided into training, validation, and testing sets following a 7:1:2 ratio for model development. Figure~\ref{fig:change-data-sentence-length-dist} illustrates the probability distribution of sentence lengths in the change category for each split. According to the figure, the samples in train, validation, and test sets have been similarly distributed in the experiments. 

\begin{figure}[htb]
    \centering
    \includegraphics[width=1\linewidth]{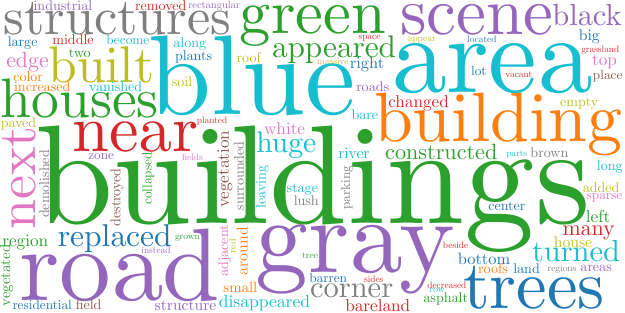}
    \caption{Word cloud based on the word frequency in the SECOND.-CC dataset. The larger the word size, the more frequently it appears in the annotated.
sentences.}
    \label{fig:wordcloud}
\end{figure}

 The word cloud of the dataset is presented in Figure~\ref{fig:wordcloud}. Some image pairs are \textit{no change} pairs, with no significant difference between the before and after images. The words, {"building", "houses", "road", "blue", "trees"} are the most commonly used words under the Change category.

\begin{figure}[hbp]
    \centering
\includegraphics[width=1\linewidth]{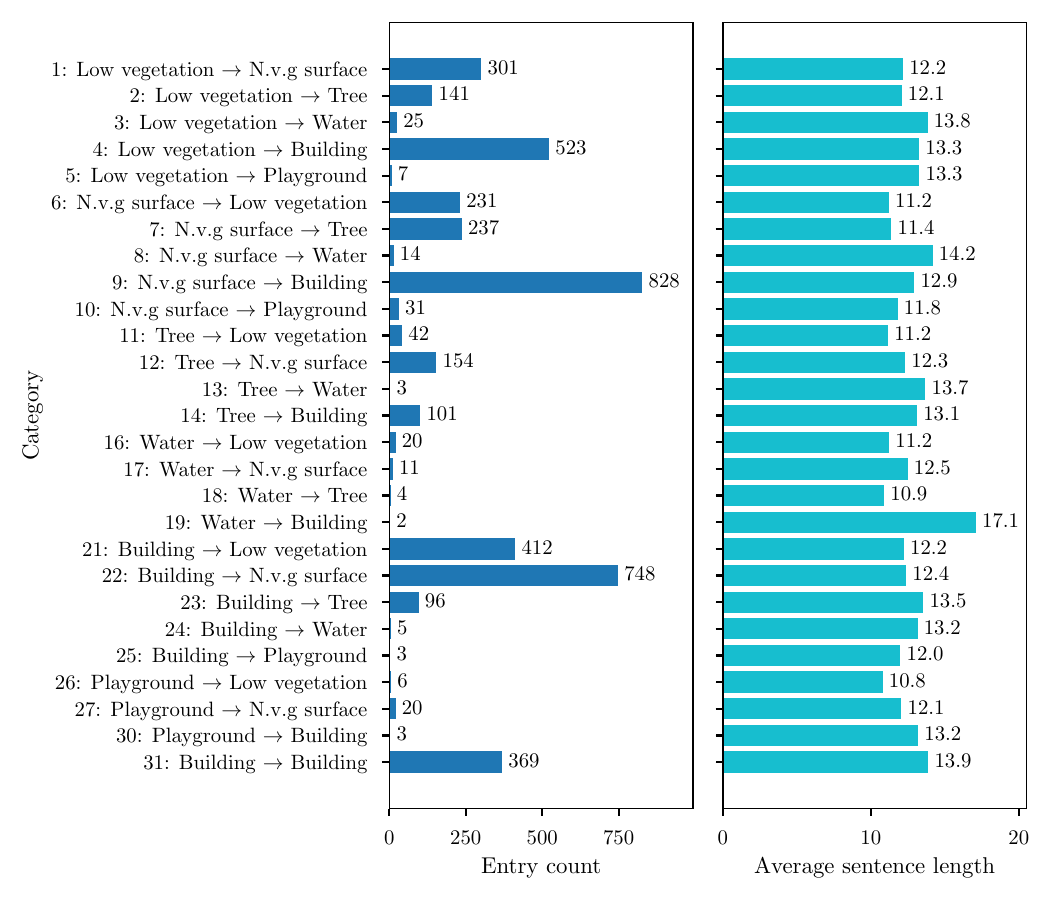}
    \caption{Bar Chart of the Distribution of Data Entries by Change Category}
    \label{fig:bar-chart-category}
\end{figure}

Last, we shared the histogram and average sentence lengths for 31 different subcategories under the change category in Figure~\ref{fig:bar-chart-category}. For example, index 1 represents that the low vegetation part of the first scene is replaced by the non-vegetation (N.v.g) surface. The number of samples in some categories, such as "N.v.g surface to building" and "building to N.v.g surface", constitutes the major part of the entire dataset due to the natural distribution of the SECOND dataset. On the other hand, some subcategories such as "water to building" have a limited number of samples compared to other subcategories. In general, average sentence lengths in subcategories seem nearly uniformly distributed except for the "water to building" subcategory with only two sample pairs.  

\subsection{RSICC Dataset Comparison}

The comparison between RSICC datasets, as shown in Table~\ref{tab:rsicc-dataset-comparison}, presents the unique properties of the SECOND-CC dataset and its augmented version in contrast to other prominent datasets such as LEVIR-CC, LEVIR-CCD, and Dubai-CCD. The proposed dataset offers several distinctive properties that make it particularly suited for advancing RSICC. First, other datasets do not contain any semantic maps whereas SECOND-CC and SECOND-CC AUG include semantic map pairs providing detailed pixel-level land-cover annotations providing additional information about the scene. Moreover, the SECOND dataset naturally incorporates spatial misalignment along with blur and brightness noise to mimic real-world remote sensing challenges. In addition, the images of the SECOND dataset have variable spatial resolutions which makes the dataset more challenging than the other dataset. Last, SECOND-CC exhibits greater variability in sentence structure, as evidenced by a higher average sentence length (10.40), a larger standard deviation (4.77) and a vocabulary size of 1\,060, which require models to process more complex syntactic and semantic patterns.

\begin{table*}[htbp]
\centering
\caption{Comparison of Remote Sensing Image Change Captioning Datasets}
\begin{tabular}{lcccccc}
\toprule
\textbf{Feature} & \textbf{LEVIR-CC} & \textbf{LEVIR-CCD} & \textbf{Dubai-CCD} & \textbf{SECOND-CC} & \textbf{SECOND-CC AUG}\\ 
\midrule
\textbf{Number of Image Pairs} & 10\,077 & 500 & 500 & 6\,041 & 10\,855\\ 
\cmidrule(lr){1-6}
\textbf{Image Size} & 256 $\times$ 256 & 256 $\times$ 256 & 50 $\times$ 50 & 256 $\times$ 256 & 256 $\times$ 256 \\ 
\cmidrule(lr){1-6}
\textbf{Number of Sentences} & 50\,385 & 2\,500 & 2\,500 & 30\,205 & 54\,275 \\ 
\cmidrule(lr){1-6}
\textbf{Vocabulary Size (words)} & 996 & 401 & 303 & 1\,060 & 1\,060\\ 
\cmidrule(lr){1-6}
\textbf{Average Sentence Length (words)} & 8.89 & 15.28 & 7.46 & 10.40 & 10.40\\ 
\cmidrule(lr){1-6}
\textbf{Standard Deviation of Sentence Length} & 4.12 & 5.69 & 3.32 & 4.77 & 4.77\\ 
\cmidrule(lr){1-6}
\textbf{Spatial Resolution} & 0.5 m/pixel & 0.5 m/pixel & 30 m/pixel & 0.5 - 3 m/pixel & 0.5 - 3 m/pixel\\ 
\cmidrule(lr){1-6}
\textbf{Spatial Misalignment} & \xmark & \xmark & \xmark & \cmark & \cmark \\
\cmidrule(lr){1-6}
\textbf{Blur \& Brightness Noise} & \xmark & \xmark & \xmark & \cmark & \cmark\\
\cmidrule(lr){1-6}
\textbf{Semantic Maps} & \xmark & \xmark & \xmark & \cmark & \cmark\\ 
\bottomrule
\end{tabular}
\label{tab:rsicc-dataset-comparison}
\end{table*}

\begin{figure}[htbp]
    \centering
    \includegraphics[width=0.95\linewidth]{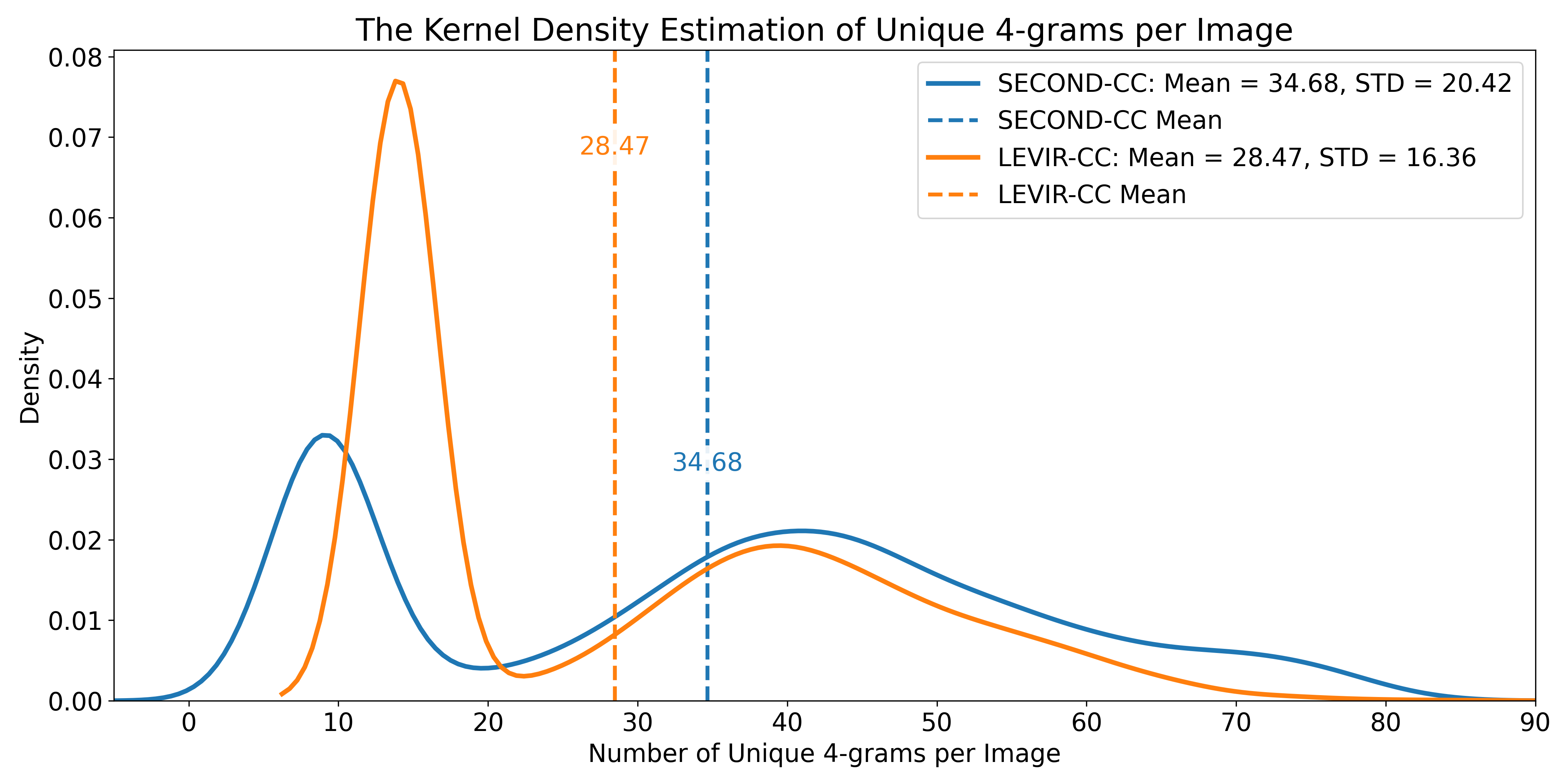}
    \caption{The Kernel Density Estimation of Unique 4-grams per Image}
    \label{fig:KDE-chart}
\end{figure}

Figure~\ref{fig:KDE-chart} provides a comparative analysis of the unique 4-gram distributions per image captions for the two larger datasets, LEVIR-CC and SECOND-CC, using kernel density estimation (KDE). This chart visualizes the probability distribution of syntactic variety in the five captions associated with each image, offering insights into the linguistic diversity promoted by each dataset. By examining the density curves, we can assess the extent to which these datasets encourage varied sentence patterns in their captions.

Specifically, SECOND-CC, compared to LEVIR-CC, shows a higher mean (on average six unique 4-grams) and broader spread of unique 4-grams compared to LEVIR-CC, indicating a richer and more varied syntactic structure in the captions per image.  Moreover, for both datasets, the number of unique 4-grams is skewed towards lower ranges. This is more apparent for the LEVIR-CC dataset, as a much higher portion of the image captions involves only 10 to 15 unique 4-grams. This suggests that the LEVIR-CC dataset tends to produce more uniform captions with less diversity. The richer syntactic variety in SECOND-CC aligns with its goal of advancing the RSICC task by pushing model capabilities to address more intricate and detailed captioning requirements.

\section{Methodology}

\begin{figure*}[hbp]
    \centering
    \includegraphics[width=0.9\linewidth]{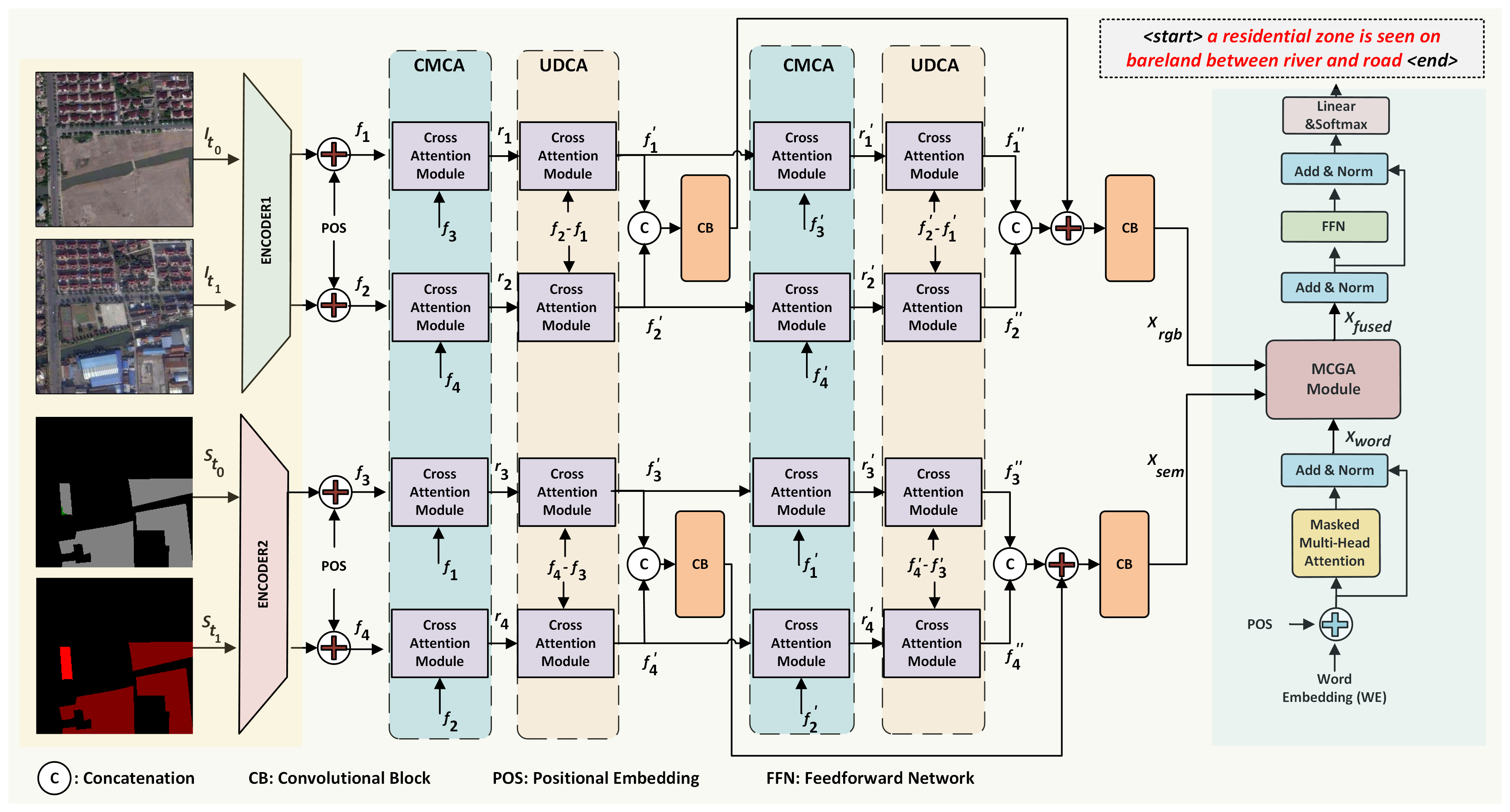}
    \caption{Overview of MModalCC, which consists of three stages: 1) an encoder, 2)  feature enhancement module with CMCA and UDCA submodules, and 3) MCGA-based caption decoder.}
    \label{fig:BlockDiagram}
\end{figure*}

The block diagram of the proposed method, MModalCC, is illustrated in Fig. \ref{fig:BlockDiagram}, and the procedure of our MModalCC model is shown in Algorithm 1. MModalCC consists of three main components: (i) an encoder, (ii) a feature enhancement module, and (iii) a caption decoder. In the training step, MModalCC takes four inputs: semantic maps and color images collected before change at $t_0$, denoted as $\mathbf{S}_{t_0}$ and $\mathbf{I}_{t_0}$ and semantic maps and color images collected after change at $t_1$, denoted as $\mathbf{S}_{t_1}$ and $\mathbf{I}_{t_1}$, respectively. First, two CNN backbones, Encoder1 and Encoder2, extract deep features from the color images $\mathbf{I}_{t_i}$ and semantic maps $\mathbf{S}_{t_i}$, respectively. Next, features are fed into a novel feature enhancement module, including a series of Cross-Modal Cross Attention (CMCA), Unimodal Difference Cross Attention (UDCA), and Convolutional Block (CB) connected with residual connections. This part efficiently fuses the semantic segmentation and RGB color image information by considering cross-modal and unimodal relations simultaneously. Finally, enhanced features of RGB and semantic images $\mathbf{x}_{rgb}$ and $\mathbf{x}_{sem}$ were applied to Multimodal Gated Cross Attention (MGCA) based caption decoder. MCGA selectively integrates multimodal features with textual context, allowing the model to focus on the most relevant visual information. This design not only considers the differences between RGB images and semantic maps but also takes into account the relationship between the images, resulting in correct language descriptions.

\subsection{Encoder}

In change captioning literature, Siamese Convolutional Neural Networks (CNNs) are widely employed for image feature extraction~\cite{CHG2CAP,RSICCformer-LEVIR-CC}. These networks ensure consistent processing of both inputs by utilizing shared weights between the twin CNNs. This design eliminates potential biases caused by differing feature extraction pathways, enabling a more robust and reliable comparison of the inputs.

Since our model processes two image pairs, as described in Algorithm \ref{alg1}, two separate Siamese backbones are required to extract features for each modality: Encoder1 and Encoder2.  For this purpose, we use the ResNet101 architecture pre-trained on the ImageNet dataset as the backbone. Specifically, the RGB image pair $(\mathbf{I}_{t_0}$, $\mathbf{I}_{t_1})$ is applied to Encoder1 whereas the semantic map image pair $(\mathbf{S}_{t_0}$ $\mathbf{S}_{t_1})$ is applied to Encoder2. By adding the positional embeddings, feature representations $({\mathbf{f}_1},{\mathbf{f}_2})$, and $({\mathbf{f}_3},{\mathbf{f}_4})$ feature pairs are obtained for RGB images and semantic maps, respectively. It is important to note that fine-tuning is applied for each encoder model, individually.
\vspace{-0.1cm}

\begin{algorithm}[htbp]

\small 
\caption{Algorithm of the proposed MModalCC Change Captioning Method}
\begin{algorithmic} 
\STATE\hspace{-0.4cm} \textbf{Input}:  Semantic and color image pairs $(\mathbf{S}_{t_0}$, $\mathbf{S}_{t_1})$ and $(\mathbf{I}_{t_0}$, $\mathbf{I}_{t_1})$.
\STATE\hspace{-0.4cm} \textbf{Output}:  \textit{Cap} $\leftarrow$Captions
\STATE\hspace{-0.4cm} \textbf{Define}: \textit{CD} $\leftarrow$ Caption Decoder,  \textit{WE} $\leftarrow$ Word Embedding, \\ \textit{CB} $\leftarrow$ Convolutional Block, \\ \textit{Encoder1} $\leftarrow$ Encoder model for RGB images, \\ \textit{Encoder2} $\leftarrow$ Encoder model for semantic maps, \\ \textit{CMCA} $\leftarrow$ Cross-Modal Cross Attention module \\ \textit{UDCA} $\leftarrow$ Unimodal Difference Cross Attention module.
\STATE\hspace{-0.4cm} -----------------------------------------------------------------------
\STATE\hspace{-0.3cm}\textbf{// Step 1}: Extract features for RGB \& SEM images and add position embeddings
\STATE\hspace{-0.3cm}1:  $\textbf{for }  i   \textnormal{ in }  {(0,1)} \textbf{ } \mathbf{do }$
\vspace{0.05cm}
\STATE\hspace{-0.3cm}2: \hspace{0.2cm} $\mathbf{f}_{i+1}   = \textit{Encoder1}( \mathbf{I}_{t_{i}}$)  + \textit{POS} 
\vspace{0.05cm}
\STATE\hspace{-0.3cm}3:\hspace{0.3cm} $\mathbf{f}_{i+3}   = \textit{Encoder2}( \mathbf{S}_{t_i}$) + \textit{POS} 
\vspace{0.05cm}
\STATE\hspace{-0.3cm}4:  $\textbf{end for}$ 
\vspace{0.05cm}
\STATE\hspace{-0.3cm}\textbf{// Step 2}: Feed the features into CMCA module 
\vspace{0.05cm}
\STATE\hspace{-0.3cm}5: [$\mathbf{r}_1, \mathbf{r}_2, \mathbf{r}_3, \mathbf{r}_4$]= \textit{CMCA}($\mathbf{f}_{1}, \mathbf{f}_{2}, \mathbf{f}_{3}, \mathbf{f}_{4}$)
\vspace{0.1cm}
\STATE\hspace{-0.3cm}\textbf{// Step 3}: Find the differences \& apply features to UDCA module 
\vspace{0.05cm}
\STATE\hspace{-0.3cm}6: $\textit{dif}(\mathbf{f}_1,\mathbf{f}_2) = \mathbf{f}_{2}-\mathbf{f}_{1}$
\vspace{0.08cm}
\STATE\hspace{-0.3cm}7: $\textit{dif}(\mathbf{f}_3,\mathbf{f}_4) = \mathbf{f}_{4}-\mathbf{f}_{3}$
\vspace{0.08cm}
\STATE\hspace{-0.3cm}8: $[{\mathbf{f}_1}', {\mathbf{f}_2}', {\mathbf{f}_3}', {\mathbf{f}_4}']$= \textit{UDCA}$(\mathbf{r}_1, \mathbf{r}_2, \mathbf{r}_3, \mathbf{r}_4$, $\textit{dif}(\mathbf{f}_1,\mathbf{f}_2), \textit{dif}(\mathbf{f}_3,\mathbf{f}_4))$
\vspace{0.1cm}
\STATE\hspace{-0.3cm}\textbf{// Step 4}: Apply the features to the CMCA module again
\vspace{0.05cm}
\STATE\hspace{-0.3cm}9: $[{\mathbf{r}_1}', {\mathbf{r}_2}', {\mathbf{r}_3}', {\mathbf{r}_4}']$= \textit{CMCA}$({\mathbf{f}_1}', {\mathbf{f}_2}', {\mathbf{f}_3}', {\mathbf{f}_4}')$
\vspace{0.1cm}
\STATE\hspace{-0.3cm}\textbf{// Step 5}: Find the differences \& apply features to UDCA module
\vspace{0.05cm}
\STATE\hspace{-0.3cm}10: $\textit{dif}({\mathbf{f}_1}',{\mathbf{f}_2}') = {\mathbf{f}_2}'-{\mathbf{f}_1}'$
\vspace{0.05cm}
\STATE\hspace{-0.3cm}11: $\textit{dif}({\mathbf{f}_3}',{\mathbf{f}_4}') = {\mathbf{f}_4}'-{\mathbf{f}_3}'$
\vspace{0.05cm}
\STATE\hspace{-0.3cm}12: $[{\mathbf{f}_1}'', {\mathbf{f}_2}'', {\mathbf{f}_3}'',{\mathbf{f}_4}'']$= $\textit{UDCA}({\mathbf{r}_1}'$, ${\mathbf{r}_2}'$,... 
\\ \hspace{4.3cm}  ${\mathbf{r}_3}', {\mathbf{r}_4}',\textit{dif}({\mathbf{f}_1}',{\mathbf{f}_2}'), \textit{dif}({\mathbf{f}_3}',{\mathbf{f}_4}'))$
\vspace{-0.05cm}
\STATE\hspace{-0.3cm}\textbf{// Step 6}: Apply the features to Convolutional Block with residual connections
\vspace{0.05cm}
\STATE\hspace{-0.3cm}13: $\mathbf{x}_{rgb} = \textit{CB}([{\mathbf{f}_1}';{ \mathbf{f}_2}'])$, $\mathbf{x}_{sem} = \textit{CB}([{\textit{ f}_3}';{ \textit{ f}_4}'])$
\STATE\hspace{-0.3cm}14: $\mathbf{x}_{rgb}$ = $\textit{CB}([{\mathbf{f}_1}'';{ \mathbf{f}_2}'']$ + $\mathbf{x}_{rgb}), \mathbf{x}_{sem} = \textit{CB}([{\mathbf{f}_3}''; {\mathbf{f}_4}'']$ + $\mathbf{x}_{sem})$
\vspace{0.1cm}
\STATE\hspace{-0.3cm}\textbf{// Step 7}: Generate captions using RGB and semantic features
\vspace{0.05cm}
\STATE\hspace{-0.3cm}15:  $\textit{Cap} = \textit{WE}(<$\textnormal{start}$>)  $
\vspace{0.05cm}
\STATE\hspace{-0.3cm}16:  $\textbf{while }  w\neq \textit{WE}(<$\textnormal{end}$>)$ do
\vspace{0.05cm}
\STATE\hspace{-0.3cm}17: \hspace{0.2cm} $ \textit{w} = \textit{CD}(x_{rgb},x_{sem},\textit{Cap})$
\vspace{0.05cm}
\STATE\hspace{-0.3cm}18: \hspace{0.2cm} $ \textit{Cap} = [\textit{Cap}; \textit{w}]$
\vspace{0.05cm}
\STATE\hspace{-0.3cm}19:  $\textbf{end while}$ 
\vspace{0.05cm}
\STATE\hspace{-0.3cm}20:  $\textbf{return} \textit{ Cap}$ 

\end{algorithmic}
\label{alg1}
\end{algorithm}

\subsection{Feature Enhancement}

Features obtained by encoders are processed by a Cross-Modal Cross-Attention (CMCA) module, which exploits the interaction between modalities to capture spatial and semantic relationships. Differences between temporal features are then computed and refined through a Unimodal Difference Cross-Attention (UDCA) module, enhancing the discriminative ability of the features. This process is iterated twice, alternating between CMCA and UDCA modules to refine and align the features further. In the subsequent step, the refined features are passed through a Convolutional Block (CB) with residual connections to consolidate RGB and semantic information while preserving feature hierarchies. Further details about each modules are provided in the subsections.

\subsubsection{Cross Attention Module}

\begin{figure}[htbp]
    \centering
    \includegraphics[width=1\linewidth]{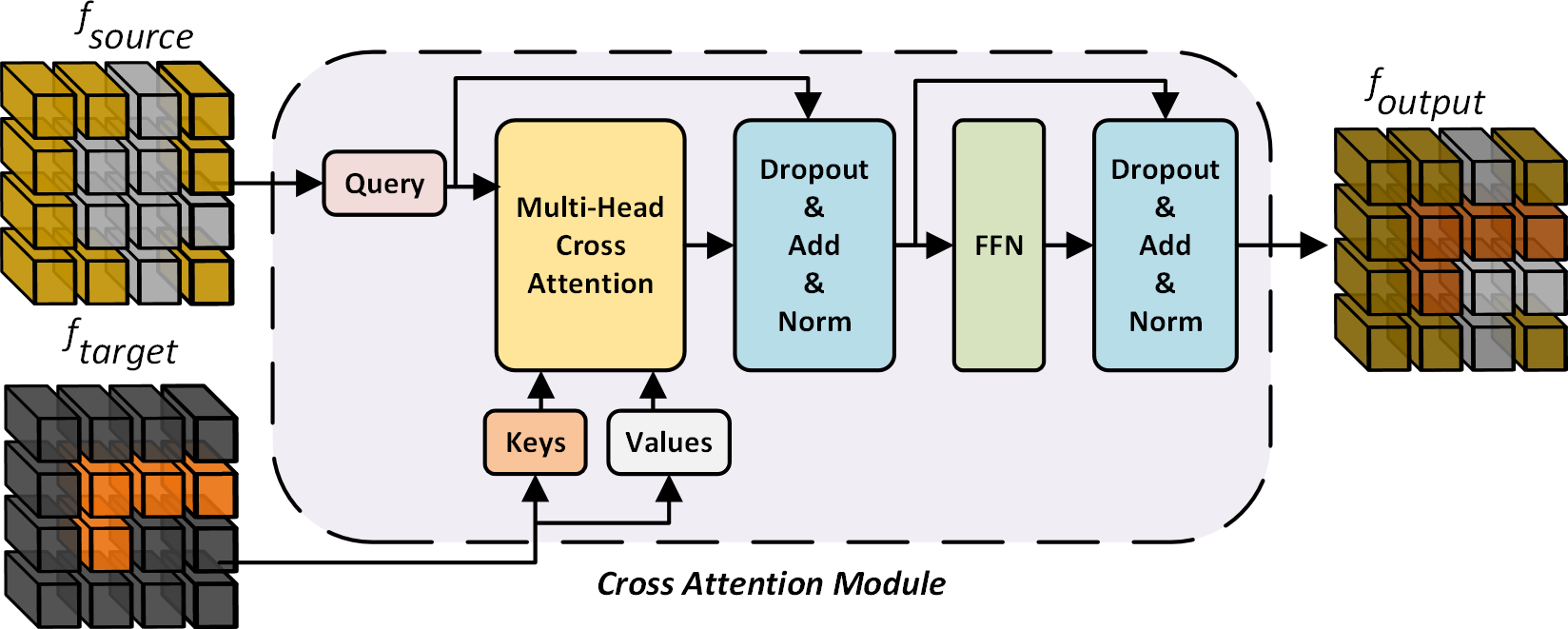}
    \caption{Structure of Cross Attention Module.}
    \label{fig:CAtt}
\end{figure}

Cross-attention is a mechanism that enables the query of the source feature $\mathbf{f}_{source}$ to focus on relevant parts of the key and queries in the target source $\mathbf{f}_{target}$, as illustrated in Fig. \ref{fig:CAtt}. It computes attention weights by comparing the query with the key and then uses them to combine the corresponding value features, highlighting important information. This allows the model to learn dependencies between two different modalities, making it particularly useful in tasks like multimodal learning such as our problem. In our research, both Cross-Modal Cross Attention (CMCA) and Unimodal Difference cross-attention (UDCA) modules use Cross Attention Modules as shown in Fig. \ref{fig:BlockDiagram}. The main difference between CMCA and UDCA has occurred in the determination of source feature $\mathbf{f}_{source}$ and target feature $\mathbf{f}_{target}$.  Using the same module for CMCA and UDCA also enables parameter sharing which provides better optimization in the training step. Therefore, the parameters of the Cross Attention Module are shared in CMCA and UDCA.

Fig. \ref{fig:CAtt} illustrates Cross Attention Module which takes the source feature $\mathbf{f}_{source} \in {\mathbb{R}}^{W^2 \times D}$  and target feature $\mathbf{f}_{target}\in {\mathbb{R}}^{W^2 \times D}$ as inputs resulting in the output feature $\mathbf{f}_{output}\in {\mathbb{R}}^{W^2 \times D}$. Here, $W^2$ denotes dimensions of grid features provided from backbone along spatial axis, while $D$ represents the number of channels in the grid features. The query $\mathbf{Q}$ ,  key $\mathbf{K}$ , and value $\mathbf{V}$ can be expressed as follows:
\begin{equation}
\mathbf{Q} = \mathbf{f}_{source} \mathbf{W}_Q,  \mathbf{K} = \mathbf{f}_{target} \mathbf{W}_K,  \mathbf{V} = \mathbf{f}_{target} \mathbf{W}_V
\end{equation}
where \(\mathbf{W}_Q, \mathbf{W}_K, \mathbf{W}_V \in \mathbb{R}^{D \times D_k}\) are learnable weight matrices and $D_k$ denote the dimension of the query, key, and value representations. Here, CMCA selects source-target feature pairs as cross-modal feature representations, \textit{i.e.}, $\mathbf{f}_{source}$ and  $\mathbf{f}_{target}$ corresponds to the feature maps of the color and semantic image before the change, respectively. On the other hand,  UDCA considers the elementwise differences as a target feature $\mathbf{f}_{target}$ in the related modality. 

The cross-attention output $\mathbf{Z}$ is computed as: 
\begin{equation}
\begin{aligned}
\begin{gathered}
\text{Attention}(\mathbf{Q}, \mathbf{K}, \mathbf{V}) = \text{softmax}\left(\frac{\mathbf{Q} \mathbf{K}^\top}{\sqrt{D_k}}\right) \mathbf{V} \\
\mathbf{Z} = \text{Attention}(\mathbf{Q}, \mathbf{K}, \mathbf{V})
\end{gathered}
\end{aligned}
\end{equation}
For multi-head attention, h parallel attention heads are used. Each head computes attention using its own set of weight matrices $\mathbf{W}_Q^i,\mathbf{W}_K^i$, and $\mathbf{W}_K^i$ for \textit{i}=1, 2, \dots, h. The outputs of all heads are concatenated and linearly transformed and cross attention output of each head is calculated as follows:
\begin{equation}
\begin{aligned}
\begin{gathered}
\mathbf{Q}_i = \mathbf{f}_s \mathbf{W}_Q^i, \quad \mathbf{K}_i = \mathbf{f}_t \mathbf{W}_K^i, \quad \mathbf{V}_i = \mathbf{f}_t \mathbf{W}_V^i  \\
\mathbf{Z}_i = \text{softmax}\left(\frac{\mathbf{Q}_i \mathbf{K}_i^\top}{\sqrt{d_k}}\right) \mathbf{V}_i
\end{gathered}
\end{aligned}
\end{equation}

Finally, the multi-head output is $\mathbf{Z}_{\text{multi-head}}$$\in {\mathbb{R}}^{W^2 \times D}$ :
\begin{equation}
\begin{split}
\mathbf{Z}_{\text{multi-head}} & = \text{MHA}(\mathbf{f}_{source},\mathbf{f}_{target},\mathbf{f}_{target}) \\ & =  \text{Concat}(\mathbf{Z}_1, \mathbf{Z}_2, \ldots, \mathbf{Z}_h) \mathbf{W}_O
\end{split}
\end{equation}
where \(\mathbf{W}_O \in \mathbb{R}^{h \cdot D_k \times D}\) is a learnable output weight matrix. Similar to traditional Transformer block, dropout, residual connections, layer normalization (LN), and feed-forward networks (FFNs) are combined sequentially after the cross-attention mechanism. It prevents vanishing or exploding gradients during training while enabling the effective capture of contextual and positional features and enhances model stability as well \cite{vaswani2017attention}. Here, the FFN consists of two fully connected linear layers, with a ReLU activation function applied between them, as described below:

\begin{equation}
\mathbf{FFN(Z)} = \text{max}(0,\mathbf{Z}\mathbf{W}_{L1})\mathbf{W}_{L2}
\end{equation}

where \(\mathbf{W}_{L1} \in \mathbb{R}^{D \times 4D}\)  and $\mathbf{W}_{L2} \in \mathbb{R}^{4D \times D}$ are learnable  weight matrices.

\subsubsection{Convolutional Block}

\begin{figure}[htbp]
    \centering
    \includegraphics[width=1\linewidth]{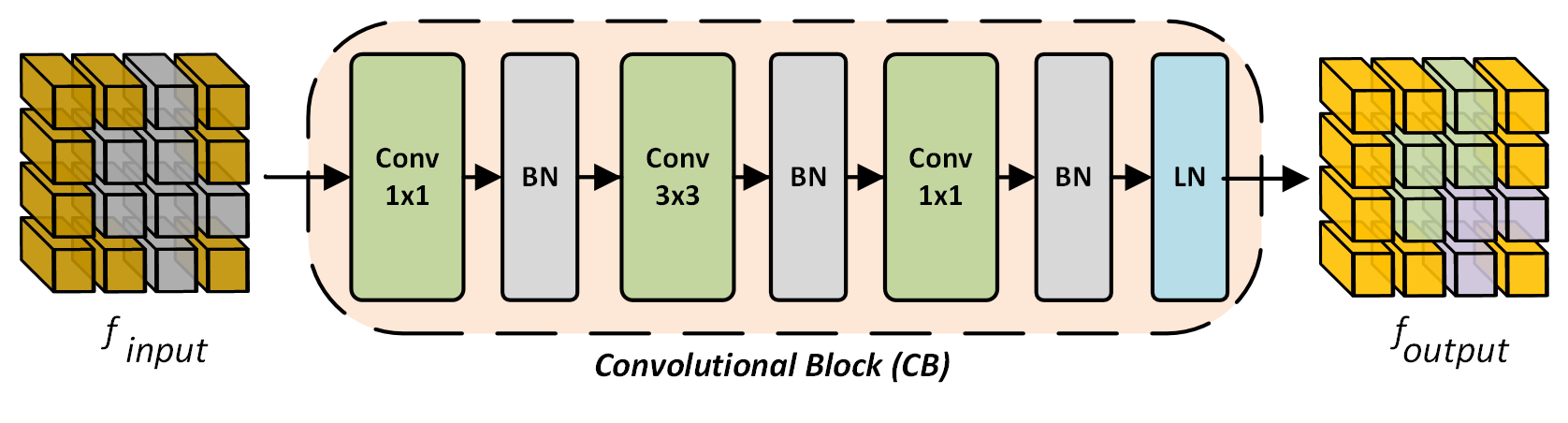}
    \caption{Structure of Convolutional Block (CB) Module.}
    \label{fig:ResBlock}
\end{figure}

The Convolutional Block (CB) module, shown in Fig. \ref{fig:ResBlock}, consists of a sequence of convolutional layers, batch normalization, and activation functions, providing feature enhancement for each modality separately. Specifically, CB includes a 1$\times$1 convolution to adjust the number of channels, followed by a 3$\times$3 convolution to capture broader spatial relationships, and another 1x1 convolution to refine the output. This process enhances the feature representations by learning both local and global patterns. 

As is seen from 6$^{th}$ step of Algorithm \ \ref{alg1},  the RGB features ($\mathbf{f}_{1}$ and $\mathbf{f}_{2}$) and semantic features ($\mathbf{f}_{3}$ and $\mathbf{f}_{4}$) are channel-wise concatenated and passed through the CB module separately. In the subsequent step, residual connections are introduced by adding the output of the CB module to the previous feature representations. Due to the channel-wise concatenation process, the output feature map is doubled in dimension axis as $\mathbf{f}_{output}\in {\mathbb{R}}^{W^2 \times 2D}$ from  $\mathbf{f}_{output}=\textit{CB}([{\mathbf{f}_1}';{ \mathbf{f}_2}']$. Therefore, the feature dimensions of the color image output $\mathbf{x}_{RGB}$ and semantic image output $\mathbf{x}_{SEM}$ at the end of the feature enhancement part, respectively.

Residual connections between CB help maintain information from earlier layers and facilitate faster convergence, enabling the network to learn more effectively. The combination of convolutional operations and residual connections before the caption decoder ensures improved performance and stability during the learning process. 

\vspace{-0.3cm}
\subsection{Multimodal Gated Cross Attention-based Decoder }

The proposed caption decoder is a transformer-based architecture that effectively integrates multi-modal information, including RGB features and semantic features, to generate descriptive captions. The decoder utilizes a Multimodal Gated Cross Attention (MGCA) mechanism to selectively fuse information from multiple modalities, followed by normalization, FFN, linear, and softmax layers for word-by-word prediction. This section describes the method with equations and detailed explanations.

\begin{figure}[ht]
    \centering
    \includegraphics[width=0.6\linewidth]{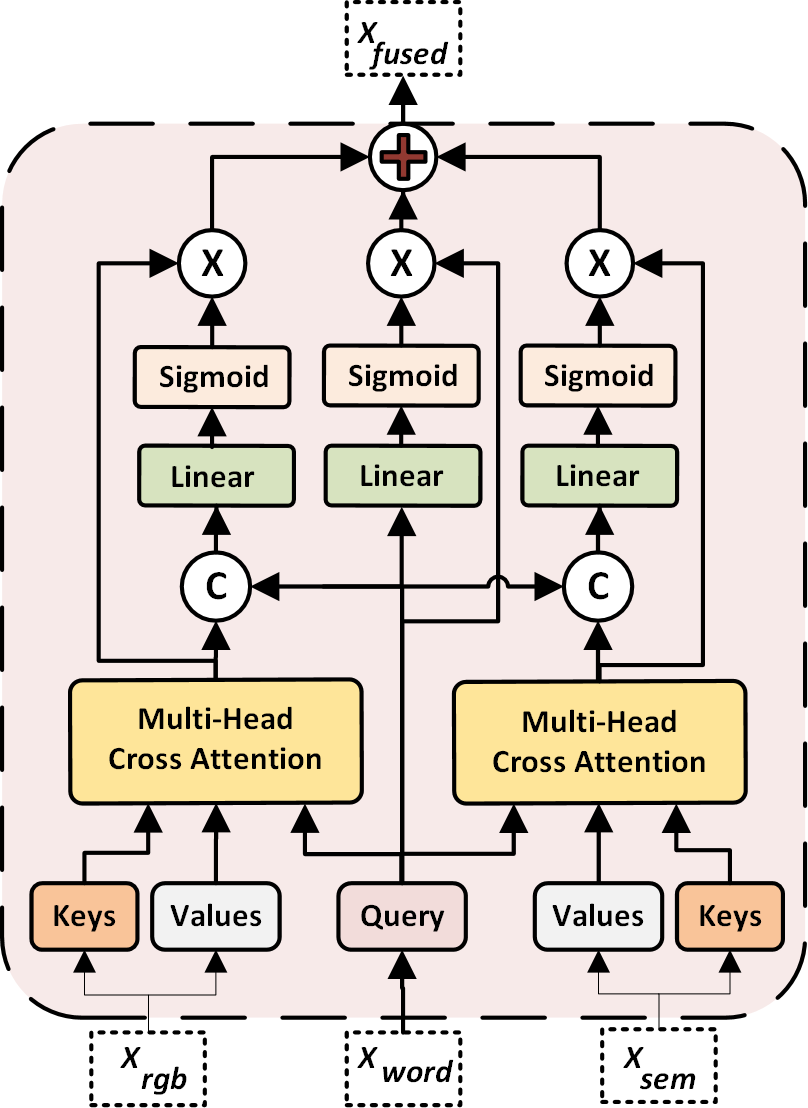}
    \caption{Multimodal Gated Cross Attention (MCGA) Module.}
    \label{fig:MCGA}
\end{figure}

As is seen from Fig. \ref{fig:BlockDiagram},  the caption decoder (CD) consists of three main stages: (i) a word embedding layer with positional encoding, (ii) an MGCA module for fusing RGB and semantic features, and (3) rest of the network to take the fused representation $\mathbf{x}_{fused}$ and produce the final prediction. The steps are detailed below.

In the first stage, the input word sequence is first mapped to an embedding vector \( \mathbf{WE} \), and positional encoding \( \mathbf{POS} \) is added to incorporate sequence order: \( \mathbf{x}_{\text{input}} = \mathbf{WE} + \mathbf{POS}\). This representation is passed through a masked multi-head attention (MHA) mechanism\cite{nicolson2020masked}.  Masked MHA enables autoregressive decoding by computing attention while masking future tokens:
\begin{equation}
\begin{aligned}
\begin{gathered}
\mathbf{Y}_i = \text{softmax}\left(\mathbf{M} + \frac{\mathbf{Q}_i \mathbf{K}_i^\top}{\sqrt{d_k}}\right) \mathbf{V}_i \\ 
\mathbf{x}_{\text{mha}} =  \text{Concat}(\mathbf{Y}_1, \mathbf{Y}_2, \ldots, \mathbf{Y}_h) \mathbf{W}_O
\end{gathered}
\end{aligned}
\end{equation}

where $\mathbf{Q}_i = \mathbf{x}_{input} \mathbf{W}_Q^i,  \mathbf{K}_i = \mathbf{x}_{input}  \mathbf{W}_K^i$, and $\mathbf{V}_i = \mathbf{x}_{input}  \mathbf{W}_V^i$. Similar to Eq. (below), each head computes attention using its own set of weight matrices $\mathbf{W}_Q^i,\mathbf{W}_K^i$, and $\mathbf{W}_K^i$ for \textit{i}=1,2,…,h. The matrix $\mathbf{M} \in \mathbb{R}^{D \times D}$ is employed to mask out similarities involving future frames, thereby maintaining causality. Since the subsequent operation is the softmax function, masking is achieved by adding $-\infty$. After that, the outputs are aggregated and passed through residual connection and normalization:
\begin{equation}
    \mathbf{x}_{\text{word}} = \text{Norm}(\mathbf{x}_{\text{mha}} + \mathbf{x}_{\text{input}}).
\end{equation}
In the second stage, MCGA, illustrated in Fig. \ref{fig:MCGA} fuses \( \mathbf{x}_{\text{word}} \), RGB features (\( \mathbf{x}_{\text{rgb}} \)), and semantic features (\( \mathbf{x}_{\text{sem}} \)). Two cross-attention heads compute attention using \( \mathbf{x}_{\text{word}} \) as follows:
\begin{equation}
\begin{aligned}
\begin{gathered}
    \mathbf{x'}_{\text{rgb}} = \text{MHA}(\mathbf{x}_{\text{word}}, \mathbf{x}_{\text{rgb}}, \mathbf{x}_{\text{rgb}}), \\
    \mathbf{x'}_{\text{sem}} = \text{MHA}(\mathbf{x}_{\text{word}}, \mathbf{x}_{\text{sem}}, \mathbf{x}_{\text{sem}}).
\end{gathered}
\end{aligned}
\end{equation}
Here, gate weights $\mathbf{g}_{\text{rgb}}$,   $\mathbf{g}_{\text{sem}}$, and $\mathbf{g}_{\text{word}}$ are calculated to determine the contribution of \( \mathbf{x}_{\text{word}} \), \( \mathbf{x'}_{\text{rgb}} \), and \( \mathbf{x'}_{\text{sem}} \) to the fused representation:
\begin{equation}
\begin{aligned}
\begin{gathered}
    \mathbf{g}_{\text{word}} = \sigma(\mathbf{W}_{\text{word}} \mathbf{x}_{\text{word}}), \\
    \mathbf{g}_{\text{rgb}} = \sigma(\mathbf{W}_{\text{rgb}} [\mathbf{x'}_{\text{rgb}} ; \mathbf{x}_{\text{word}}]), \\
    \mathbf{g}_{\text{sem}} = \sigma(\mathbf{W}_{\text{sem}} [\mathbf{x'}_{\text{sem}} ; \mathbf{x}_{\text{word}}]),
\end{gathered}
\end{aligned}
\end{equation}
where $\mathbf{W}_{\text{sem}},\mathbf{W}_{\text{rgb}}  \in {\mathbb{R}}^{4D \times 2D}$ and $\mathbf{W}_{\text{word}}\in {\mathbb{R}}^{2D \times 2D}$ are learnable parameters and \( \sigma \) is the sigmoid activation. The fused representation is computed as a weighted sum:
\begin{equation}
    \mathbf{x}_{\text{fused}} = \mathbf{g}_{\text{word}} \odot \mathbf{x}_{\text{word}} + \mathbf{g}_{\text{rgb}} \odot \mathbf{y}_{\text{rgb}} + \mathbf{g}_{\text{sem}} \odot \mathbf{y}_{\text{sem}}
\end{equation}
where \( \odot \) denotes element-wise multiplication. Finally, the fused feature \( \mathbf{z} \) is processed by the following equations:
\begin{align}
\begin{gathered}
    \mathbf{x}_{\text{final}} = \text{Norm(Norm}(\mathbf{x}_{\text{fused}}) + \text{FFN(Norm}(\mathbf{x}_{\text{fused}}))), \\
    \mathbf{p} = \text{softmax}(\mathbf{W}_{\text{out}} \mathbf{x}_{\text{final}}),
\end{gathered}
\end{align}
where \( \mathbf{p} \) represents the predicted probability distribution over the vocabulary.

\section{Experimental Results}


This section evaluates the MModalCC method through a series of experiments designed to explore its strengths and analyze its performance across different scenarios. The implementation details, including training configurations and evaluation metrics are introduced. The impact of data augmentation is examined along with the key ablation studies that explore the contributions of certain architectural components, such as the effect of multi-modal cross-attention and different decoder configurations, supported by visual representations of encoder and decoder attention maps. Finally, a comprehensive performance comparison table is provided, showing the advantages of MModalCC over baseline and state-of-the-art methods, and demonstrating how the proposed approach effectively addresses the challenges of remote sensing change captioning tasks. It is important to note that experiments in "Ablation Study" and "Performance Comparison" subsections are carried out by using augmented version of the dataset, SECOND-CC AUG, due to its effectiveness shared in "Normal Data versus Augmented Data" subsection.

\subsection{Experimental Setup}

We developed our models with PyTorch framework. All models are trained and results are evaluated on the NVIDIA GTX 4090 GPU with 24GB VRAM. In the training stage, we use the cross-entropy loss function and Adam\cite{Adam} optimizer with $5\times10^{-5}$ learning rate. The maximum epoch is set to 30. The vision layer we train in our model is ResNet-101\cite{Resnet-101} convolution layer trained on ImageNet\cite{ImageNet} dataset.

Selecting the capable captioning metrics is essential to establish an effective and meaningful comparison of the performance of visual interpretation models. Following \cite{GlobalVisualFeatureandLignusticState}, the metrics are chosen as BLEU-N (N-1,2,3,4)\cite{BLEU}, ROUGE$_L$\cite{ROUGE}, METEOR\cite{METEOR}, CIDEr-D\cite{CIDEr}, SPICE\cite{SPICE} and an average metric, denoted as $S_m^*$, to evaluate the quality of the generated captions. These metrics collectively provide a framework for evaluating the quality of captions with their ability to measure syntactic accuracy, contextual coherence and semantic relevance of the generated captions to the reference sentences. The CIDEr-D metric is not considered for no change results as this method penalize frequent words. Since the words in the no change reference captions are identical, the CIDEr-D metric assigns a score of 0 to every generated sentence in this category, thus losing its capability to provide meaningful scores. Thus, CIDEr-D metric is omitted in these calculations and $S_m^*$ is calculated using the remaining four metrics.

\begin{equation}
S_m^*  = \dfrac{(B4+METEOR+ROUGE+CIDEr+SPICE)}{5} 
\end{equation}

\subsection{Beam Size Search}

Beam search is a search algorithm used to generate sequences by exploring multiple possible options at each step and selecting the most likely ones. Here, beam size determines how many of these options are kept for further exploration during the captioning. Through the experiments conducted with varying beam size hyperparameters, we observed that the performance of the MModalCC model improved as the beam size increased. Specifically, the model achieved a slightly higher $S_m^*$ score when the beam size reached 4, which was the optimal value in our validation tests. This suggests that a beam size of 4 balances exploration and computational efficiency, leading to better performance. Smaller beam sizes, on the other hand, may limit the model’s ability to explore diverse captioning possibilities. In comparison, larger beam sizes could introduce unnecessary computational overhead without a significant improvement in the quality of the results.

\begin{table}[htbp]
\centering
\caption{Evaluation Metrics for Different Beam Sizes}
\label{tab:evaluation_metrics_beam_sizes}
\resizebox{0.5\textwidth}{!}{
\begin{tabular}{r|cccccc}
\toprule
\textbf{Beam Size}  & \textbf{BLEU4} & \textbf{METEOR} & \textbf{ROUGE} & \textbf{CIDEr} & \textbf{SPICE} & \textbf{$S_m^*$} \\
\midrule
1  & $0.374$ & $\mathbf{0.283}$ & $0.579$ & $0.925$ & $\mathbf{0.258}$ & $0.484$ \\
2 & $0.379$ & $0.281$ & $0.584$ & $\mathbf{0.933}$ & $0.252$ & $0.486$ \\
3 & $0.381$ & $0.280$ & $\mathbf{0.585}$ & $0.930$ & $0.249$ & $0.485$ \\
4 & $\mathbf{0.386}$ & $0.280$ & $0.584$ & $\mathbf{0.933}$ & $0.249$ & $\mathbf{0.487}$ \\
5 & $0.385$ & $0.279$ & $0.585$ & $0.932$ & $0.248$ & $0.486$ \\
\bottomrule
\end{tabular}
}
\end{table}

\subsection{Normal Data versus Augmented Data}

Considering the various challenges present in SECOND-CC, which involves change captioning across remote sensing images, we found it essential to employ data augmentation techniques. These challenges include image registration errors, brightness differences, blurriness, and variations in viewing angles, all of which can significantly hinder the model’s ability to effectively detect and describe changes. 

We explored various data augmentation procedures during training to enhance the diversity of our dataset. In our augmented dataset, we applied one of these techniques randomly to each image, generating transformed images and doubling the dataset size as described in Section 2. 

Table \ref{tab:comparison_aug_vs_no_aug} shows the comparison of MModalCC performances in SECOND-CC and SECOND-CC AUG datasets. Here, performance metrics are measured not only entire dataset but also change and no-change categories to interpret the results better. According to table, there is a significant improvement in overall scores after augmentation in SECOND-CC AUG dataset. Augmented dataset led to greater increase in scores at change images compared to others. The augmentation also resulted with meaningfull improvement in overall score indicating its positive effect on the models general understanding. 

In summary, this approach allowed us to better train the model under more diverse conditions, leading to improved performance and greater robustness in change captioning tasks. To this end, all subsequent experiments were conducted using the SECOND-CC AUG. 

\begin{table*}[t]
\centering
\caption{Comparison of MModalCC Performance with and without Data Augmentation}
\label{tab:comparison_aug_vs_no_aug}
\begin{tabular}{lcccccccccc}
\toprule
\textbf{Category} & \textbf{Augmentation} & \textbf{BLEU1} & \textbf{BLEU2} & \textbf{BLEU3} & \textbf{BLEU4} & \textbf{METEOR} & \textbf{ROUGE} & \textbf{CIDEr} & \textbf{SPICE} & \textbf{$S_m^*$} \\
\midrule
\multirow{2}{*}{Overall} & \cmark & \textbf{0.697} & \textbf{0.562} & \textbf{0.461} & \textbf{0.386} & \textbf{0.280} & \textbf{0.584} & \textbf{0.933} & \textbf{0.249} & \textbf{0.487} \\
& \xmark & 0.687 & 0.556 & 0.456 & 0.383 & 0.274 & 0.574 & 0.898 & 0.242 & 0.474 \\
\midrule
\multirow{2}{*}{Change} & \cmark & \textbf{0.636} & \textbf{0.472} & \textbf{0.349} & \textbf{0.261} & \textbf{0.220} & \textbf{0.435} & \textbf{0.512} & \textbf{0.186} & \textbf{0.323} \\
& \xmark & 0.619 & 0.456 & 0.330 & 0.240 & 0.210 & 0.424 & 0.469 & 0.171 & 0.303 \\
\midrule
\multirow{2}{*}{No-Change} & \cmark & \textbf{0.957} & \textbf{0.949} & \textbf{0.944} & \textbf{0.940} & \textbf{0.735} & \textbf{0.972} & - & 0.413 & \textbf{0.612} \\
&\xmark& 0.950 & 0.940 & 0.933 & 0.927 & 0.718 & 0.964 & - & \textbf{0.426} & 0.607 \\
\bottomrule
\end{tabular}
\end{table*}

\subsection{Ablation Study}

\subsubsection{The effect of Multi-Modal Cross-Attention}
\label{sssec:multimodalCA}
This ablation study examines which configurations of cross-attention modules, detailed in Figure \ref{fig:CAtt}, are most effective and advantageous for integrating RGB images and semantic maps. The performance is evaluated across various metrics under these configurations:

\begin{enumerate}
    \item Single-Modality Cross-Attention: Cross-attention applied only to RGB features (CMCA) or semantic features (UDCA).
    \item Dual Cross-Attention: Cross-attention applied bidirectionally, allowing RGB features to attend to semantic features and semantic features to attend to RGB features.
\end{enumerate}

The results of these configurations with and without cross-attention for each modality are shown in Table \ref{tab:ablation_cattention}. 

Results show that, in most cases, employing dual cross-attention (both CMCA and UDCA) achieves a higher performance across change and no-change cases. 

For change cases, the dual-attention configuration consistently improves the metrics, increasing $S_m^*$ significantly with respect to other settings, showing its ability to capture complementary information from RGB and semantic modalities. However, for no-change scenes, the impact of dual cross-attention is less pronounced. Interestingly, single-modality cross-attention configurations (particularly CMCA) demonstrate slightly better scores on certain metrics, such as SPICE and BLEU. Otherwise, the dual cross-attention configuration scores are comparable to the scores of single-modality configurations.

In the overall scores, the inclusion of dual cross-attention augments the model's capability to interpret change scenes, increasing $S_m^*$ from $0.462$ (CMCA only) and $0.455$ (UDCA only) to $0.487$, while maintaining decent performance in scenes with no-change. This suggests that cross-modal integration with CMCA and UDCA proves beneficial for handling the complex contextual relations between features in change captioning tasks.

\begin{figure*}[htbp]
    \centering
    \includegraphics[width=1\linewidth]{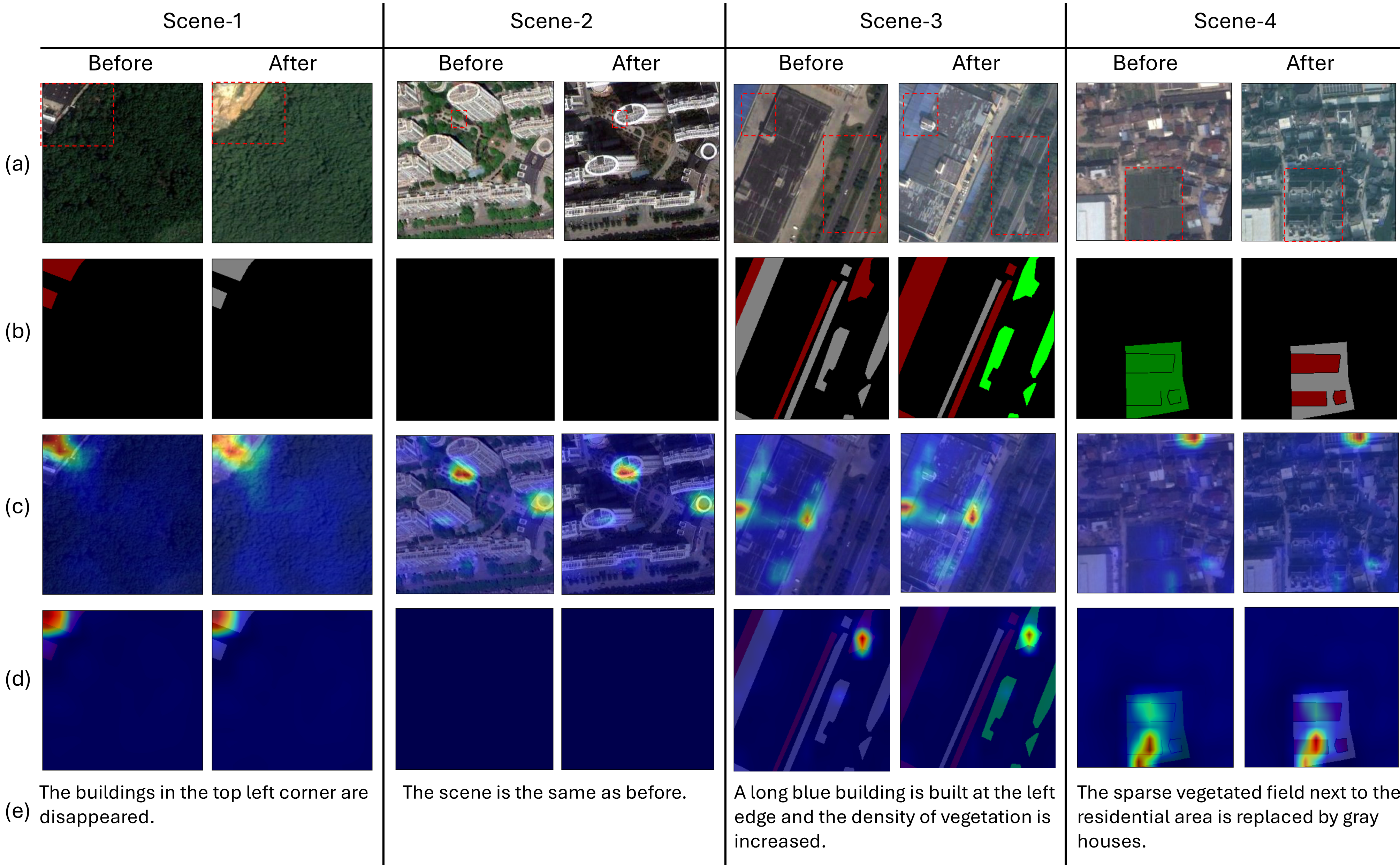}
    \caption{Encoder Attention Maps. Illustrates four scenes, with Scene-2 being a no-change case and the others showing changes. For each scene, row (a) displays before and after RGB images with the changes seen in the RGB versions highlighted with red rectangles. Row (b) shows semantic maps, row (c) RGB encoder attention maps, and row (d) semantic encoder attention maps. The captions generated for the before and after image pairs are shown in row (e).}
    \label{fig:EncAttMap}
\end{figure*}

To further illustrate these findings, Figure \ref{fig:EncAttMap} presents several examples of change and no-change scenes, with the changed areas highlighted using red rectangles on the RGB images. In certain cases, such as the no-change example in Scene-2, variations in capturing angles cause structures in the scene to appear tilted or changed. Detecting this as a no-change scenario solely using RGB features (Fig. \ref{fig:EncAttMap}a) poses a significant challenge, as indicated by the encoder attention maps (Fig. \ref{fig:EncAttMap}c), which incorrectly detect changes due to these visual distortions. However, the inclusion of semantic context helps clarify the scene by providing correcting categorical and spatial information. For instance, in Scene-2 (b), the semantic map indicates no changes between temporal states, appearing as an entirely black map. By incorporating cross-attention between RGB and semantic features, MModalCC effectively integrates this information, as shown in the zero-attentive map in Scene 2 (d) and in the generated caption, which accurately reflects the absence of changes.

In complex change scenarios, the RGB encoder attention maps sometimes wrongly focus on areas unrelated to actual changes, leading to misrepresentation. However, the integration of semantic maps provides essential categorical and spatial context, rectifying the error by using the distilled information provided by semantic maps, as shown in the corrected semantic encoder attention maps. By making use of the dual cross-attention, MModalCC effectively integrates RGB and semantic features to improve change captioning both contextually and semantically, as evidenced by the accuracy and relevance of the generated captions (e.g., Scenes 3 and Scenes 4).

The results of this ablation study show that employing a dual cross-attention mechanism overall augments the model to interpret changes effectively. Bidirectional information flow between RGB and semantic feature learning allows better use of each feature set. This reinforces our decision to adopt dual cross-attention with CMCA and UDCA as a core component of our model framework.


\begin{table*}[t]
\centering

\caption{Ablation Studies on the Different Encoder Attention Configurations}
\begin{tabular}{lcc|ccccccccc}
\toprule
\textbf{Category} & \textbf{CMCA} & \textbf{UDCA} & \textbf{BLEU1} & \textbf{BLEU2} & \textbf{BLEU3} & \textbf{BLEU4} & \textbf{METEOR} & \textbf{ROUGE} & \textbf{CIDEr} & \textbf{SPICE} & \textbf{$S_m^*$} \\
\midrule
\multirow{3}{*}{Overall} 
&\xmark& \cmark & 0.647 & 0.520 & 0.429 & 0.361 & 0.263 & 0.556 & 0.859 & 0.234 & 0.455 \\
& \cmark &\xmark& 0.663 & 0.535 & 0.434 & 0.358 & 0.262 & 0.569 & 0.864 & \textbf{0.257} & 0.462 \\
& \cmark & \cmark & \textbf{0.697} & \textbf{0.562} & \textbf{0.461} & \textbf{0.386} & \textbf{0.280} & \textbf{0.584} & \textbf{0.933} & 0.249 & \textbf{0.487} \\
\midrule
\multirow{3}{*}{Change} 
&\xmark& \cmark & 0.573 & 0.416 & 0.302 & 0.219 & 0.198 & 0.401 & 0.425 & 0.158 & 0.280 \\
& \cmark &\xmark& 0.595 & 0.440 & 0.315 & 0.224 & 0.201 & 0.414 & 0.415 & 0.167 & 0.284 \\
& \cmark & \cmark & \textbf{0.636} & \textbf{0.472} & \textbf{0.349} & \textbf{0.261} & \textbf{0.220} & \textbf{0.435} & \textbf{0.512} & \textbf{0.186} & \textbf{0.323} \\
\midrule
\multirow{3}{*}{No-Change} 
&\xmark& \cmark & 0.941 & 0.931 & 0.924 & 0.919 & \textbf{0.710} & 0.959 & \textbf{-} & 0.433 & 0.755 \\
& \cmark &\xmark& \textbf{0.961} & \textbf{0.953} & \textbf{0.949} & \textbf{0.945} & 0.707 & 0.961 & \textbf{-} & \textbf{0.470} & \textbf{0.771} \\
& \cmark & \cmark & 0.957 & 0.949 & 0.944 & 0.940 & 0.735 & \textbf{0.972} & \textbf{-} & 0.413 & 0.765 \\
\bottomrule
\label{tab:ablation_cattention}
\end{tabular}
\end{table*}

\begin{figure*}[htbp]
    \centering
    \includegraphics[width=1\linewidth]{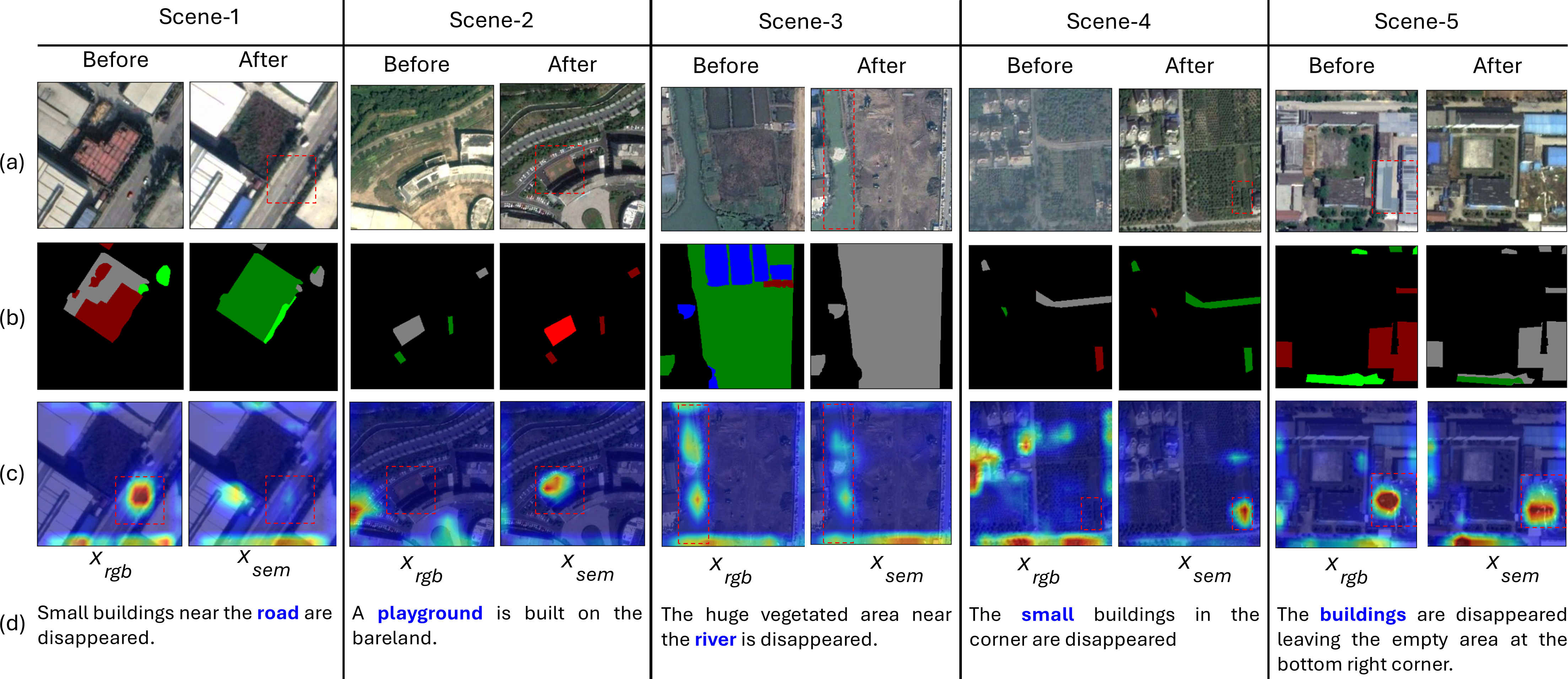}
    \caption{Decoder Attention Maps. Illustrates results under the dual-modality configuration for five scenes, with all scenes corresponding to change-cases. The scenes show the distinct capabilities of decoder attention, where the RGB and semantic components often complement or correct each other by focusing on specific objects and changes in the images. For each scene; row (a) displays before and after RGB images with the object of interest highlighted with a red rectangle, row (b) shows the semantic maps for each temporal instance, row (c) provides decoder attention maps independently for $\mathbf{x}_{rgb}$ and $\mathbf{x}_{sem}$, highlighting the regions of focus based on RGB and semantic change features, overlaid as heatmaps on the after images, row (d) shows the generated captions, with bold blue words highlighting the regions of change aligned with the $\mathbf{x}_{rgb}$ and $\mathbf{x}_{sem}$ attention maps.}
    \label{fig:decoderAtt}
\end{figure*}


\subsection{The Effect of Decoder Configurations}

This ablation study examines the impact of different decoder configurations, specifically the use of RGB ($\mathbf{x}_{rgb}$) and semantic ($\mathbf{x}_{sem}$) features, in the Multimodal Gated Cross Attention-based decoder. To fully exploit the complementary nature of RGB and semantic features, we retain both CMCA and UDCA modules active throughout this study. This configuration allows the decoder to integrate bidirectional cross-modal attention, providing an effective flow of learning exchange between the two modalities, as discussed in Section \ref{sssec:multimodalCA}. The performance is evaluated across various metrics under three configurations:

\begin{enumerate}
    \item Semantic-only ($\mathbf{x}_{sem}$): Only semantic features are used.
    \item RGB-only ($\mathbf{x}_{rgb}$): Only RGB features are used.
    \item Dual-modality ($\mathbf{x}_{rgb}~\&~\mathbf{x}_{sem}$): Both RGB and semantic features are combined.
\end{enumerate}

In change cases, the dual-modality configuration shows notable improvements in capturing changes. This indicates a successfully established complementary relationship between semantic and RGB features in identifying and describing changes in the scenes. The most significant improvements with dual-modality enabled are observed in BLEU4 and CIDEr metrics, which emphasize capturing precise contextual alignment with the objects of the scene. Single-modality configurations struggle in change scenarios, falling short of achieving the same level of contextual detail and, consequently, lower performance in these metrics.

In no-change cases, all configurations perform comparably, with marginal differences in BLEU and $S_m^*$ scores. The dual-modality setup achieves the highest $S_m^*$ score, which is closely followed by the $\mathbf{x}_{sem}$ configuration. Notably, the semantic-only approach yields slightly higher BLEU scores than the RGB-only configuration in no-change cases, indicating a potential susceptibility in the RGB feature space where minor pixel-level variations could be misinterpreted as changes.

In the overall scope, the dual-modality configuration, where both $\mathbf{x}_{rgb}$ and $\mathbf{x}_{sem}$ features included, consistently achieves highest scores compared to any single-modality setup across all metrics. The dual-modality configuration ($\mathbf{x}_{rgb}~\&~\mathbf{x}_{sem}$) demonstrates the benefits of combining RGB and semantic features for handling complex changes and improving overall object-context relationship in the captions, as can be deducted from significantly higher BLEU4 and CIDEr scores of this configuration. 

To further evaluate the benefits of the dual-modality configuration, some change-case samples with decoder attention maps with respect to either of the $\mathbf{x}_{rgb}$ and $\mathbf{x}_{sem}$ features are depicted along with the generated captions in Figure \ref{fig:decoderAtt}. These examples demonstrate how RGB and semantic features complement each other in representing change cases. For instance, the attention maps highlight cases where one modality focuses on the correct object or region (e.g., Scene 1 and Scene 2), while the other may focus elsewhere. This complementary behavior ensures that critical details are not missed, even when one feature set diverges.

In addition, the dual-modality approach often results in improved overall attention to the correct regions, as can be observed in cases where both feature sets align (e.g., Scene 3 and Scene 5). This alignment enhances semantic coherence and contributes to more accurate caption generation. In other instances, when one modality lacks precision, the other effectively compensates, such as when $\mathbf{x}{sem}$ correctly identifies smaller or less prominent objects while $\mathbf{x}{rgb}$ attends to broader regions (e.g., Scene 4).

The findings from this ablation study, reinforced by both performance metrics and attention map analyses, strongly support the choice of the dual-modality configuration in decoder setups. Furthermore, the attention maps demonstrate a complementary relationship between the two modalities: when one modality falls short in focusing on the correct region, the other often compensates effectively, as highlighted in Scene 1 and Scene 2 in Figure~\ref{fig:decoderAtt}.

\begin{table*}[t]
\centering
\caption{Ablation Studies on the Different Decoder Configurations}
\begin{tabular}{lcc|ccccccccc}
\toprule
\textbf{Category} & \textbf{$\mathbf{x}_{rgb}$} & \textbf{$\mathbf{x}_{sem}$} & \textbf{BLEU1} & \textbf{BLEU2} & \textbf{BLEU3} & \textbf{BLEU4} & \textbf{METEOR} & \textbf{ROUGE} & \textbf{CIDEr} & \textbf{SPICE} & \textbf{$S_m^*$} \\
\midrule
\multirow{3}{*}{Overall} 
& \xmark & \cmark & 0.676 & 0.542 & 0.438 & 0.362 & 0.266 & 0.568 & 0.869 & 0.232 & 0.459 \\
& \cmark & \xmark & 0.667 & 0.534 & 0.433 & 0.356 & 0.262 & 0.562 & 0.831 & 0.242 & 0.451 \\
& \cmark & \cmark & \textbf{0.697} & \textbf{0.562} & \textbf{0.461} & \textbf{0.386} & \textbf{0.280} & \textbf{0.584} & \textbf{0.933} & \textbf{0.249} & \textbf{0.487} \\
\midrule
\multirow{3}{*}{Change} 
& \xmark & \cmark & 0.610 & 0.446 & 0.316 & 0.224 & 0.204 & 0.413 & 0.413 & 0.162 & 0.283 \\
& \cmark & \xmark & 0.603 & 0.442 & 0.317 & 0.224 & 0.203 & 0.414 & 0.394 & 0.156 & 0.278 \\
& \cmark & \cmark & \textbf{0.636} & \textbf{0.472} & \textbf{0.349} & \textbf{0.261} & \textbf{0.220} & \textbf{0.435} & \textbf{0.512} & \textbf{0.186} & \textbf{0.323} \\
\midrule
\multirow{3}{*}{No-Change} 
& \xmark & \cmark & \textbf{0.959} & \textbf{0.951} & \textbf{0.946} & \textbf{0.942} & 0.732 & 0.969 & \textbf{-} & 0.413 & 0.764 \\
& \cmark & \xmark & 0.921 & 0.907 & 0.897 & 0.888 & 0.686 & 0.946 & \textbf{-} & \textbf{0.466} & 0.747 \\
& \cmark & \cmark & 0.957 & 0.949 & 0.944 & 0.940 & \textbf{0.735} & \textbf{0.972} & \textbf{-} & 0.413 & \textbf{0.765} \\
\bottomrule
\label{tab:ablationdecoder}
\end{tabular}
\end{table*}

\vspace{-0.2cm}

\subsection{Performance Comparison}

\begin{figure*}
    \centering
    \includegraphics[width=0.98\linewidth]{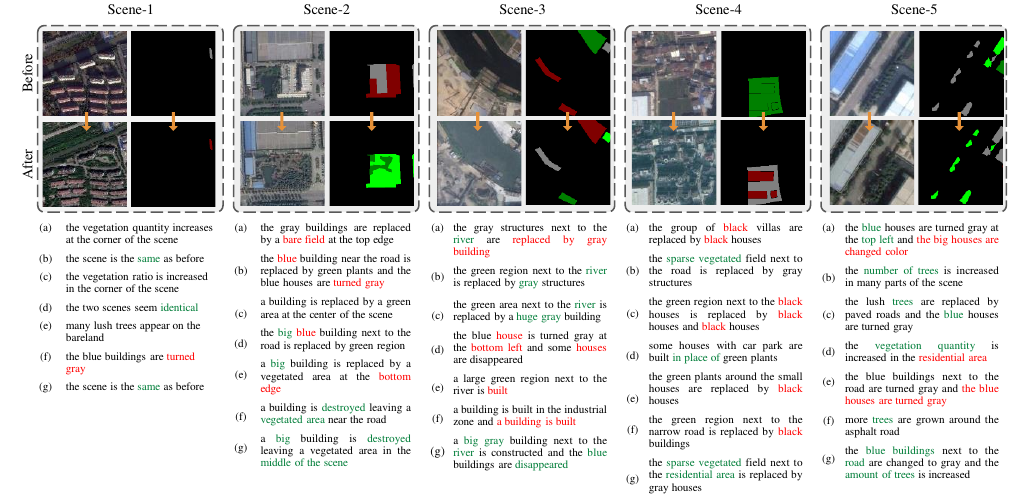}
    \caption{Comparison of captions generated by different RSICC models for examples from the SECOND-CC AUG dataset. Each scene is shown with a pair of RS images and corresponding semantic maps, along with captions generated by the following models: (a) Chg2Cap trained on RGB images, (b) Chg2Cap trained on semantic maps, (c) PSNet trained on RGB images, (d) PSNet trained on semantic maps, (e) RSICCformer trained on RGB images, (f) RSICCformer trained on semantic maps, and (g) MModalCC. Parts of captions that contain grammatical or semantic errors are marked in red, while those that provide additional correct information compared to others are highlighted in green.}
    \label{fig:model-comparison}
\end{figure*}

A performance comparison is conducted among state-of-the-art benchmark frameworks, including RSICCformer, Chg2Cap, and PSNet, using both RGB and semantic (SEM) versions of the proposed SECOND-CC AUG dataset. Since these frameworks are not inherently designed to integrate multi-modal data, the dataset is provided in separate RGB and SEM formats for fine-tuning within their existing architectures. 

Table~\ref{tab:method_comparison} shows the performance comparison of these models over the metrics and Figure~\ref{fig:model-comparison} compares the trained models over captions generated for the given RGB and semantic context. The results explain the significant advantages of the proposed MModalCC framework over existing methods.

According to the table, across all evaluation metrics, MModalCC achieves superior scores, clearly outperforming RGB and SEM variants of RSICCformer, Chg2Cap, and PSNet. Using MModalCC provides substantial gains in captioning performance when compared to the next best-performing framework. Among the competing frameworks, PSNet-RGB and RSICCformer-RGB scores 2nd and 3rd highest scores across all metrics, respectively. Specifically, for the $S_m^*$ metric, MModalCC achieves 0.487, surpassing the next best scores of 0.447 (Chg2Cap-RGB) and 0.438 (RSICCformer-RGB) by approximately 9\%. Similarly, for the BLEU4 metric, MModalCC outperforms Chg2Cap-RGB by a notable margin of 13.20\%, and for the CIDEr metric, it achieves an 11.48\% improvement. 

The consistency of MModalCC's performance can be further validated by the scene descriptions presented in Figure~\ref{fig:model-comparison}. This figure compares captions generated by various RSICC models for pairs of RS images and their corresponding semantic maps with the generated captions, across five diverse scenes. The figure focuses on cornerstones of change captioning such as capturing relative spatial directions, accurately describing pre- and post-change states with precise language, and maintaining contextual relevance along with logical consistency.

\begin{table*}[t]
\centering
\caption{Performance Comparison of Change Captioning Methods}
\begin{tabular}{l|cccccccccc}
\toprule
\textbf{Method} & \textbf{BLEU1} & \textbf{BLEU2} & \textbf{BLEU3} & \textbf{BLEU4} & \textbf{METEOR} & \textbf{ROUGE} & \textbf{CIDEr} & \textbf{SPICE} & \textbf{$S_m^*$} \\
\midrule
RSICCformer-RGB & 0.665 & 0.518 & 0.415 & 0.340 & 0.262 & 0.547 & 0.810 & 0.232 & 0.438 \\
Chg2Cap-RGB & 0.670 & 0.521 & 0.416 & 0.341 & 0.266 & 0.551 & 0.837 & 0.238 & 0.447 \\
PSNET-RGB & 0.663 & 0.506 & 0.398 & 0.322 & 0.263 & 0.547 & 0.787 & 0.237 & 0.431 \\
RSICCformer-SEM & 0.641 & 0.494 & 0.396 & 0.330 & 0.255 & 0.540 & 0.786 & 0.228 & 0.428 \\
Chg2Cap-SEM & 0.610 & 0.454 & 0.359 & 0.295 & 0.250 & 0.523 & 0.743 & 0.221 & 0.406 \\
PSNET-SEM & 0.634 & 0.476 & 0.378 & 0.312 & 0.250 & 0.528 & 0.751 & 0.217 & 0.412 \\
MModalCC & \textbf{0.697} & \textbf{0.562} & \textbf{0.461} & \textbf{0.386} & \textbf{0.280} & \textbf{0.584} & \textbf{0.933} & \textbf{0.249} & \textbf{0.487} \\
\bottomrule
\end{tabular}
\label{tab:method_comparison}
\end{table*}

MModalCC stands out for its reliability in identifying the absence of change, a distinction shared only by Chg2Cap \& PSNet frameworks trained on semantic maps in this example (Scene-1). The added value of both semantic and RGB features, with CMCA and UDCA, further increases the accuracy of its descriptions as well as its positional precision. 

In change cases, MModalCC demonstrates its strength in accurately describing spatial relationships, both in terms of relative directional phrases (in relation to other objects in the scene) and event positioning within the scene (e.g., "in the middle of the scene", "next to the river", "next to the residential area") (Scenes 2-4). Other models sometimes misrepresent or omit these critical details. 

MModalCC minimizes issues like redundancy in descriptions, where other models sometimes generate repetitive phrases by repeating object names or characteristics within a single caption decoding (as seen in Scene-3 (f), Scene-4 (a), Scene-4 (c), and Scene 5 (e)).

A captioning model must generate descriptions with logical consistency by accurately using object-context relationships within a scene. This requires aligning actions and transformations with the inherent properties of objects, avoiding illogical statements such as "building a green region" (e.g. Scene-3 (f)). For example, logical consistency ensures that man-made structures are appropriately described as "constructed" or "demolished", while natural elements are correctly represented as "transformed", "replaced" or "disappeared". MModalCC achieves this logical consistency through its MGCA decoder. The MGCA selectively fuses RGB and semantic features with joined context, providing relevant details from each modality for generating logically sound sentences with strong object-context relationships.


\section{Conclusion}

In this paper, we addressed the critical challenges in remote sensing change captioning, such as illumination differences, viewpoint variations, blur effects, and image registration errors. To tackle these issues, we introduced SECOND-CC, a publicly available dataset that provides 6041 high-resolution RGB image pairs and semantic segmentation maps at mixed spatial resolutions (0.5-3 m/pixel). Detailed change descriptions offer a comprehensive benchmark for RSICC research. Furthermore, we also prepared an augmented dataset SECOND-CC AUG with 10\,855 image pairs and 54\,275 change descriptions to reach better performances in this challenging dataset. 

We also proposed MModalCC, a novel multimodal framework that effectively integrates semantic and visual data through advanced attention mechanisms. Extensive experiments demonstrated that MModalCC significantly outperforms existing methods, including RSICCformer, Chg2Cap, and PSNet with +4.6\% improvement on BLEU4, +9.6\% improvement on CIDEr scores, and +3.1\% improvement on ROUGE scores. The results of detailed ablation studies and attention map analyses validated the effectiveness of our design steps and provided insights into the contributions of individual components.

This study establishes a strong foundation for future RSICC research by introducing a robust dataset, proposing an effective multimodal method, and offering actionable insights for model design. Future work will explore additional modalities and expand the dataset to cover even more diverse and challenging scenarios.

This study developed the MModalCC model using pairs of semantic and RGB images provided by the SECOND-CC dataset. However, the estimation of semantic maps from RGB images is a well-established approach in the literature and has been successfully demonstrated using various deep-learning models. Studies leveraging the SECOND dataset often achieve promising results in generating semantic maps with a certain level of accuracy~\cite{chen2024changemamba,ding2024joint}. Nevertheless, the inherent errors in these predicted semantic maps can influence change captioning performance negatively. This aspect has not been addressed within the scope of this paper but will be a focus of our future work. In this regard, a deeper investigation into the impact of segmentation errors on change summarization and an analysis of the model's adaptability and robustness could provide broader applicability for the proposed approach.

\vspace{-0.15cm}
\section*{Acknowledgments}
This research was supported by the Scientific and Technological Research Council of Turkey (TUBITAK) under grant 3501, project number 122E666. We would like to thank the authors of the SECOND dataset for providing a publicly available change detection dataset. We are also grateful to the authors of RSICCformer, Chg2Cap, and PSNet papers.
\vspace{-0.15cm}

\end{document}